\documentclass{article} 
\usepackage{iclr2027_conference,times}

\usepackage{amsmath,amsfonts,bm}

\def\eqref#1{equation~\ref{#1}}

\def\1{\bm{1}}

\DeclareMathAlphabet{\mathsfit}{\encodingdefault}{\sfdefault}{m}{sl}
\SetMathAlphabet{\mathsfit}{bold}{\encodingdefault}{\sfdefault}{bx}{n}

\DeclareMathOperator*{\argmin}{arg\,min}

\usepackage{hyperref}
\usepackage{url}

\usepackage[utf8]{inputenc}
\usepackage[T1]{fontenc}
\usepackage[english]{babel}
\usepackage{csquotes}
\usepackage[dvipsnames]{xcolor}

\usepackage{amsmath,amsfonts,amssymb,mathtools}

\usepackage{amsthm}
\usepackage{thmtools}
\usepackage{thm-restate}

\usepackage{graphicx}
\usepackage{pgfplots}
\usepackage{pgfplotstable}
\pgfplotsset{compat=1.18}
\usepackage{algorithm}
\usepackage{algorithmic}
\usepackage{booktabs}
\usepackage{array}
\usepackage{multirow}
\usepackage{makecell}
\usepackage{colortbl}
\usepackage{subcaption}
\usepackage{wrapfig}

\usepackage{tikz}
\usetikzlibrary{
    arrows.meta,
    backgrounds,
    calc,
    fit,
    positioning,
    shapes,
    shapes.geometric
}
\usepackage{tikz-cd}

\usepackage{hyperref}
\usepackage{xspace}
\usepackage{pifont}
\usepackage[capitalize,noabbrev]{cleveref}

\definecolor{sthlmBlue}{RGB}{0,110,191}
\definecolor{sthlmLightBlue}{RGB}{214,237,252}
\definecolor{sthlmGreen}{RGB}{0,134,127}
\definecolor{sthlmLightGreen}{RGB}{213,247,244}
\definecolor{sthlmRed}{RGB}{196,0,100}
\definecolor{sthlmLightRed}{RGB}{254,222,237}
\definecolor{sthlmOrange}{RGB}{221,74,44}
\definecolor{sthlmLightOrange}{RGB}{255,215,210}
\definecolor{sthlmPurple}{RGB}{93,35,125}
\definecolor{sthlmLightPurple}{RGB}{241,230,252}
\definecolor{sthlmYellow}{RGB}{252,191,10}
\definecolor{sthlmDarkGrey}{RGB}{51,51,51}
\definecolor{sthlmGrey}{RGB}{245,243,238}
\definecolor{sthlmLightGrey}{RGB}{213,217,225}

\definecolor{terrainBase}{RGB}{238,240,238}
\definecolor{terrainHigh}{RGB}{225,230,225}
\definecolor{asphalt}{RGB}{65,70,75}

\definecolor{scrollBlue}{HTML}{1F5AA6}
\definecolor{scrollRed}{HTML}{C43C39}
\definecolor{scrollPurple}{HTML}{6A3FB5}
\definecolor{scrollInk}{HTML}{2B2B2B}
\definecolor{scrollMuted}{HTML}{7A7A7A}
\definecolor{scrollPanel}{HTML}{F7F7F7}
\definecolor{scrollLine}{HTML}{D9D9D9}

\hypersetup{
    colorlinks=true,
    linkcolor=sthlmRed,
    citecolor=sthlmGreen,
    urlcolor=sthlmBlue
}

\newcommand{\AddPlayerAndTimeStep}[3]{
    \ifblank{#2}{\ifblank{#3}{#1}{{#1}_{#3}}}{\ifblank{#3}{{#1}^{#2}}{{#1}^{#2}_{#3}}}
}

\newcommand{\DualGap}[1]{
  \ifblank{#1}{d}{d_{#1}}
}
\newcommand{\PessimisticValue}[2]{
    \AddPlayerAndTimeStep{\underline{v}}{#1}{#2}
}
\newcommand{\PessimisticQValue}[2]{
    \AddPlayerAndTimeStep{\underline{q}}{#1}{#2}
}
\newcommand{\Dynamics}[2]{
    \AddPlayerAndTimeStep{F}{#1}{#2}
}

\newcommand{\Player}[1]{{#1}}

\newcommand{\EFG}[1]{
    \ifblank{#1}{\mathcal{E}}{\mathcal{E}^{#1}}
}

\newcommand{\StateSpace}{\Set{S}}
\newcommand{\State}[1]{s_{#1}}
\newcommand{\BigState}[1]{S_{#1}}

\newcommand{\ActionSpace}[2]{\AddPlayerAndTimeStep{\Set{A}}{#1}{#2}}
\newcommand{\JointActionSpace}[1]{\ActionSpace{}{#1}}

\newcommand{\Action}[2]{\AddPlayerAndTimeStep{a}{#1}{#2}}
\newcommand{\JointAction}[1]{\Action{}{#1}}

\newcommand{\BigAction}[2]{\AddPlayerAndTimeStep{A}{#1}{#2}}
\newcommand{\BigJointAction}[1]{\BigAction{}{#1}}

\newcommand{\ObservationSpace}[2]{\AddPlayerAndTimeStep{\Set{Z}}{#1}{#2}}
\newcommand{\JointObservationSpace}[1]{\ObservationSpace{}{#1}}

\newcommand{\Observation}[2]{\AddPlayerAndTimeStep{z}{#1}{#2}}
\newcommand{\JointObservation}[1]{\Observation{}{#1}}

\newcommand{\BigObservation}[2]{\AddPlayerAndTimeStep{Z}{#1}{#2}}

\newcommand{\HistorySpace}[2]{\AddPlayerAndTimeStep{\Set{H}}{#1}{#2}}
\newcommand{\JointHistorySpace}[1]{\HistorySpace{}{#1}}

\newcommand{\PublicHistorySpace}[1]{\HistorySpace{0}{#1}}

\newcommand{\History}[2]{\AddPlayerAndTimeStep{h}{#1}{#2}}
\newcommand{\JointHistory}[1]{\History{}{#1}}

\newcommand{\PublicHistory}[1]{\History{0}{#1}}

\newcommand{\BigHistory}[2]{\AddPlayerAndTimeStep{H}{#1}{#2}}
\newcommand{\BigJointHistory}[1]{\BigHistory{}{#1}}

\newcommand{\BigPublicHistory}[1]{\BigHistory{0}{#1}}

\newcommand{\PolicySpace}[2]{\AddPlayerAndTimeStep{\Pi}{#1}{#2}}
\newcommand{\JointPolicySpace}[1]{\PolicySpace{}{#1}}
\newcommand{\PrefixPolicySpace}[2]{\PolicySpace{#1}{0:#2}}
\newcommand{\JointPrefixPolicySpace}[1]{\PolicySpace{}{0:#1}}
\newcommand{\SuffixPolicySpace}[2]{\PolicySpace{}{#2:\Horizon}}
\newcommand{\JointSuffixPolicySpace}[1]{\PolicySpace{}{#1:\Horizon}}

\newcommand{\Policy}[2]{\AddPlayerAndTimeStep{\pi}{#1}{#2}}
\newcommand{\JointPolicy}[1]{\Policy{}{#1}}
\newcommand{\PrefixPolicy}[2]{\Policy{#1}{0:#2}}
\newcommand{\JointPrefixPolicy}[1]{\Policy{}{0:#1}}
\newcommand{\SuffixPolicy}[2]{\Policy{#1}{#2:\Horizon}}
\newcommand{\JointSuffixPolicy}[1]{\Policy{}{#1:\Horizon}}

\newcommand{\PublicObservationSpace}[1]{\ObservationSpace{0}{#1}}

\newcommand{\PublicObservation}[1]{\AddPlayerAndTimeStep{z}{0}{#1}}

\newcommand{\BigPublicObservation}[1]{\AddPlayerAndTimeStep{Z}{0}{#1}}

\newcommand{\RewardFunction}[1]{\AddPlayerAndTimeStep{r}{#1}{}}

\newcommand{\TransitionFunction}{p}

\newcommand{\BeliefState}[2]{b^{#1}_{#2}}
\newcommand{\BeliefSpace}[2]{\Set{B}^{#1}_{#2}}

\newcommand{\ConditionalState}[2]{c^{#1}_{#2}}
\newcommand{\BigConditionalState}[2]{C^{#1}_{#2}}
\newcommand{\ConditionalSpace}[2]{\Set{C}^{#1}_{#2}}

\newcommand{\ContVal}[2]{\ValueFunction^{\mathrm{cont}}_{#1,#2}}

\newif\ifdetails
 \detailsfalse     

\newcommand{\Set}[1]{\mathcal{#1}}

\newcommand{\Ex}{\mathbb{E}}

\newcommand{\Prob}{\mathbb{P}}
\newcommand{\Simplex}[1]{\Delta(#1)}
\newcommand{\RealSpace}{\mathbb{R}}
\newcommand{\defeq}{\doteq}

\newcommand{\ie}{\emph{i.e.}}

\newcommand{\cf}{\emph{cf.}}

\newcommand{\LocalGameTuple}[2]{\Set{M}^{#1}_{#2}}
\newcommand{\FocalBeliefGameTuple}[1]{\Set{M}^{#1}}
\newcommand{\GameTuple}{\Set{M}}

\newcommand{\Agents}{\Set{I}}

\newcommand{\Horizon}{T}

\newcommand{\InitialBelief}{b_0}

\newcommand{\TransitionCondKernel}[2]{\TransitionFunction^{\mathtt{c}}_{#1,#2}}

\newcommand{\AlgConDecRule}[3]{\bar{\pi}^{#3}_{#1,#2}}
\newcommand{\AlgExpDecRule}[3]{\tilde{\pi}^{#3}_{#1,#2}}

\newcommand{\PrivPolicyValue}[2]{J^{#2}_{#1}}

\newcommand{\PrivValFunc}[3]{\ValueFunction^{#3}_{#1,#2}}
\newcommand{\OptBeliefVal}[2]{\underline{\ValueFunction}^{#1, *}_{#2}}
\newcommand{\BeliefQVal}[2]{\hat{\QFn}_{#1,#2}}

\newcommand{\OptimalPrivateValueFunction}[2]{\ValueFunction^{#1,*}_{#2}}

\newcommand{\ValueFunction}{v}
\newcommand{\ValueFunctionErwan}{\underline{v}}
\newcommand{\QFn}{q}

\newcommand{\RedTrans}[1]{\tau_{#1}}
\newcommand{\RedRew}[1]{\rho_{#1}}
\newcommand{\RetBeliefRew}[2]{g^{#1}_{#2}}
\newcommand{\RetCondRew}[2]{g^{\mathtt{c}}_{#1,#2}}

\newcommand{\RetBeliefTrans}[2]{\tau^{#1}_{#2}}
\newcommand{\RetCondTrans}[2]{\tau^{\mathtt{c}}_{#1,#2}}
\newcommand{\RetCondObs}[2]{w^{\mathtt{c}}_{#1,#2}}

\newcommand{\UninfState}[1]{o_{#1}}

\newcommand{\UninfSpace}[1]{\Set{O}_{#1}}

\definecolor{sthlmBlue}{RGB}{0,110,191}
\definecolor{sthlmGreen}{RGB}{0,134,127}
\definecolor{sthlmRed}{RGB}{196,0,100}
\definecolor{sthlmPurple}{RGB}{93,35,125}
\definecolor{sthlmDarkGrey}{RGB}{51,51,51}
\definecolor{PanelGray}{RGB}{245,247,250}

\theoremstyle{plain}
\newtheorem{theorem}{Theorem}
\newtheorem{proposition}{Proposition}
\newtheorem{lemma}{Lemma}

\theoremstyle{definition}
\newtheorem{definition}{Definition}

\theoremstyle{remark}



\newcommand{\red}[1]{\textcolor{sthlmRed}{#1}}

\newcommand{\opti}[1]{\textbf{\red{#1}}}

\title{Conservative Policy Iteration for Zero-Sum POSGs}

\iclrfinalcopy

\author{%
  Erwan C. Escudie\\
  \texttt{e.c.escudie@rug.nl}\\
  University of Groningen
  \And Matthia Sabatelli\\
  \texttt{m.sabatelli@rug.nl}\\
  University of Groningen
  \And Jilles Steeve Dibangoye \\
  \texttt{j.s.dibangoye@rug.nl} \\
  University of Groningen 
}

\begin{document}

\maketitle

\begin{abstract}

Occupancy-state reductions make dynamic programming possible for zero-sum partially observable stochastic games (zs-POSGs). Existing point-based solvers usually optimise a central value function. This restores Bellman recursion, but it requires point-based updates over high-dimensional occupancy spaces. We introduce an approximate policy-iteration algorithm for finite-horizon zs-POSGs. The method maintains one policy per player and updates both jointly along occupancy trajectories induced by conservative policy updates. Instead of approximating a full central value function, each update uses a single supporting hyperplane induced by the current joint policy.  The algorithm alternates local policy improvements, conservative updates, and persistent exploration. Computation is therefore concentrated on occupancy regions where the current joint policy is exploitable. On finite-horizon zero-sum POSG benchmarks, the method is competitive at small horizons and often improves the runtime--exploitability trade-off at larger horizons compared with PBVI, HSVI, and CFR+.
\end{abstract}

\section{Introduction}

Planning under imperfect information requires agents to act from histories rather than from the true state.
In partially observable Markov decision processes (POMDPs), belief states restore a Markovian dynamic program and support exact and point-based solvers \citep{astro-1965:optimcontr:journal:174,smallwood,sondik78,pineau2003point,SmithSimmons2004HSVI}.
In partially observable stochastic games (POSGs), partial observability is coupled with strategic uncertainty.
Each player must reason about the hidden state, its own information, and the possible information and actions of the other players \citep{HansenBZ04,Kovarik2022}.
The two-player zero-sum case, called a zero-sum partially observable stochastic game (zs-POSG), is the adversarial form of this problem.
Dynamic programming is classical for fully observable stochastic games \citep{shapley,filar-competitiveMDP}, but partial observability requires a sufficient statistic for planning.
Common-information, plan-time, and occupancy-state reductions provide such statistics for decentralised control and Dec-POMDPs \citep{nayyar2013decentralized,DibangoyeABC13,Dibangoye:OMDP:2016,oliehoek2013sufficient}.
Recent work has extended this programme to adversarial settings, including value-structure results, one-sided partially observable stochastic games, public-observation games, and general two-player zs-POSGs \citep{Wiggers16,horak2017heuristic,HorBos-aaai19,Nayyar2019,Buffet2020,DelBufDibSaf-DGAA-23,SokotaDL0KB23,Escudie2025}.
In these reductions, a zs-POSG is lifted to an occupancy stochastic game.
The lifted state is a distribution over hidden states and histories.
The lifted actions are decision rules.
The reduced game is fully observed by the planner.
It preserves the value of the original game and restores Bellman recursion.

This progress leaves an algorithmic gap.
Existing dynamic-programming solvers for zs-POSGs are mostly point-based value-iteration methods \citep{pineau2003point,SmithSimmons2004HSVI,DelBufDibSaf-DGAA-23,Escudie2025}.
They approximate the optimal central value on sampled occupancy states and apply Bellman backups, for all samples in sync.
This is principled, but it creates a coverage problem.
The approximation must support backups over a large occupancy space.
Player-specific policy improvement needs accurate value information only where the current policy can fail to be secure.
In zero-sum games, these regions are not determined by the current policy alone.
They are exposed by unilateral deviations and worst-case responses.
The difficulty is therefore not only to extract executable policies from a central planner.
It is the mismatch between global value approximation and local policy improvement.

Policy-update methods address another side of the same problem.
Fictitious play, counterfactual regret minimisation, and CFR+ update policies directly in imperfect-information games \citep{Robinson1951,Zinkevich2007,Neller2013AnIT,Tammelin14,Moravck2017DeepStackEA,BrownBLG20}.
Classical fictitious play, regret minimisation, and CFR-style methods show that repeated local improvement can solve large zero-sum imperfect-information games without explicitly representing a global dynamic program \citep{Robinson1951,Zinkevich2007,Neller2013AnIT,Tammelin14}.
Sequence-form methods and information-set decompositions provide compact representations of extensive-form games 
\citep{SFLP,Moravck2017DeepStackEA,BrownBLG20,Kovarik2022}.
Policy-improvement and trust-region methods provide another route to executable updates in reinforcement learning and multi-agent reinforcement learning \citep{BertsekasTsitsiklis1996,Munos2003,Kakade2002,Schulman2015,Schulman2017,Yu2022,Zhong2024}.
Search and policy improvement have also been combined in cooperative or partially observable multi-agent settings \citep{lerer2020improving,van2016coordinated,wei2019presslight,pieroth2024detecting}.
These methods are attractive because they update policies at histories or information states that are executable by construction.

Existing methods provide either a principled value equation or executable policy updates.
They do not yet provide both in a form suited to focused approximation.
The central occupancy reduction provides a Bellman equation, but its state is broad and not directly executable.
Focal planning makes policy extraction player-specific, but its state is set-valued and lacks a tractable value representation.
Policy-space methods update executable policies, but they do not provide a Bellman representation for focused point-based backups.
The missing object is therefore a player-specific value representation that keeps executable decision rules while supporting local Bellman backups.

To address this gap, we introduce Conservative Policy Iteration (CPI) for finite-horizon zs-POSGs.
CPI maintains one policy per player and improves both policies along occupancy trajectories induced by conservative policy updates.
For each player, the current policy induces a value function in the corresponding player-specific planning process.
CPI uses this policy-induced value function for local Bellman backups, instead of constructing a global central value function.
Each backup computes an improved decision rule for one player from its current value function. 
The two policies are updated together, so the opposing policy used during rollouts is supplied by the other player-specific planning process.
The improved decision rules are averaged into the current policies, and GLIE pertubations maintain exploration.
Thus, updates are performed where the current joint policy is tested by adversarial interaction, rather than by uniformly covering the central occupancy space.
The convergence guarantees are conditional on exact local backups, sufficient coverage, and the stated regularity assumptions.
We evaluate CPI on standard finite-horizon zero-sum POSG benchmarks. 
The method is competitive at small horizons.
At larger horizons, it often improves the runtime-exploitability trade-off compared with PBVI \citep{Escudie2025}, HSVI \citep{DelBufDibSaf-DGAA-23}, and CFR+ \citep{Tammelin14}.

\section{Preliminaries}

Finite-horizon zero-sum POSGs are stochastic games with hidden states and imperfect information.

\begin{definition}
A finite-horizon zero-sum partially observable stochastic game (zs-POSG) is a tuple 
\(\GameTuple=(
    \Agents,
    \StateSpace,
    \JointActionSpace,
    \JointObservationSpace,
    \PublicObservationSpace{},
    \TransitionFunction,
    (\RewardFunction{i})_{i\in\Agents},
    \Horizon,
    \InitialBelief
    )
\), where \(\Agents=\{1,2\}\), \(\StateSpace\) is a finite state space, \(\JointActionSpace{}=\times_{i\in\Agents}\ActionSpace{i}{}\) is the joint action space, \(\JointObservationSpace{}=\times_{i\in\Agents}\ObservationSpace{i}{}\) is the joint private observation space, \(\PublicObservationSpace{}\) is the public observation space, \(\TransitionFunction:\StateSpace\times\JointActionSpace{}\to\Simplex{\StateSpace\times\PublicObservationSpace{}\times\JointObservationSpace{}}\) is the transition kernel, \(\RewardFunction{i}:\StateSpace\times\JointActionSpace{}\to\RealSpace\) is player \(i\)'s reward, \(\RewardFunction{2}=-\RewardFunction{1}\), \(\InitialBelief\in\Simplex{\StateSpace}\) is the initial state distribution, and \(\Horizon\) is the horizon.
\end{definition}

The game unfolds over stages \(t=0,\ldots,\Horizon-1\). At stage \(t\), the players choose \(\JointAction{t}\), each player $\Player{i}$ receives \(\RewardFunction{i}(\State{t},\JointAction{t})\), and the game samples \((\State{t+1},\PublicObservation{t+1},\JointObservation{t+1})\) according to \(\TransitionFunction(\cdot\mid \State{t},\JointAction{t})\). Because the state is hidden, $\Player{i}$ acts from its private history \(\History{i}{t}\) and the public history \(\PublicHistory{t}\). These histories are empty at \(t=0\). For \(t\ge1\), \(\History{i}{t}\doteq \Action{i}{0}\Observation{i}{1}\cdots\Action{i}{t-1}\Observation{i}{t}\) records $\Player{i}$'s past actions and private observations, and \(\PublicHistory{t}\doteq \PublicObservation{1}\cdots\PublicObservation{t}\) records the public observations. We write \(\JointHistory{t}\defeq(\History{1}{t},\History{2}{t})\) for a joint private history, with reachable sets \(\HistorySpace{i}{t}\), \(\JointHistorySpace{t}\), and \(\PublicHistorySpace{t}\).

A decision rule for player \(i\) is a map \(\Policy{i}{t}:\HistorySpace{i}{t}\times\PublicHistorySpace{t}\to\Simplex{\ActionSpace{}{i}}\). A joint decision rule is \(\JointPolicy{t}=(\Policy{1}{t},\Policy{2}{t})\in\JointPolicySpace{t}\defeq\PolicySpace{1}{t}\times\PolicySpace{2}{t}\). A behavioural policy profile is \(\JointPolicy{}=(\JointPolicy{0},\ldots,\JointPolicy{\Horizon-1})\in\JointPolicySpace{}\). We write \(\Policy{i}{}\in\PolicySpace{i}{}\) for a full behavioural policy of player \(i\). We write \(\SuffixPolicy{i}{t}\in\SuffixPolicySpace{i}{t}\) for player \(i\)'s policy suffix and \(\JointSuffixPolicy{t}\in\JointSuffixPolicySpace{t}\) for a policy-profile suffix from stage \(t\) onward. We write \(\PrefixPolicy{i}{t}\in\PrefixPolicySpace{i}{t}\) for player \(i\)'s policy prefix and \(\JointPrefixPolicy{t}\in\JointPrefixPolicySpace{t}\) for a policy-profile prefix.

Policies are evaluated by expected cumulative payoff. Given \(\InitialBelief\), $\Player{i}$'s payoff under \(\JointPolicy{}\) is \(\PrivPolicyValue{i}{\JointPolicy{}}\defeq\Ex_{\InitialBelief,\TransitionFunction,\JointPolicy{}}[\sum_{k=0}^{\Horizon-1}\RewardFunction{i}(\BigState{k},\BigJointAction{k})]\). Since \(\RewardFunction{2}=-\RewardFunction{1}\), we write the zero-sum game from player \(1\)'s perspective. The finite-horizon zs-POSG induces a finite zero-sum extensive-form game with perfect recall, so the classical minimax and behavioural-policy foundations apply \citep{Neumann1928,kuhn1953}.
Thus,
\[
\OptimalPrivateValueFunction{1}{0}{}(\InitialBelief)
=
\max_{\Policy{1}{}\in\PolicySpace{1}{}}\min_{\Policy{2}{}\in\PolicySpace{2}{}}\PrivPolicyValue{1}{(\Policy{1}{},\Policy{2}{})}
=
\min_{\Policy{2}{}\in\PolicySpace{2}{}}\max_{\Policy{1}{}\in\PolicySpace{1}{}}\PrivPolicyValue{1}{(\Policy{1}{},\Policy{2}{})}.
\]
A policy profile attaining this value is a Nash equilibrium of the induced zero-sum game. Equivalently, player \(i\)'s component is maximin optimal for $\Player{i}$'s payoff for any \(i\in\Agents\). This player-specific view is central below. A value-based method must ultimately select decision rules that can be implemented from each player's private and public information.

\subsection{Occupancy-state dynamic programming}

Bellman recursion relies on a state that is sufficient for the continuation problem.
%
For a policy-profile prefix \(\JointPrefixPolicy{t}\), the central occupancy state is \(\UninfState{t}(\JointPrefixPolicy{t})=\Prob_{\InitialBelief,\TransitionFunction,\JointPrefixPolicy{t}}(\BigState{t},\BigJointHistory{t},\BigPublicHistory{t})\).
We write \(\UninfState{t}\) when the prefix is clear.
It belongs to \(\UninfSpace{t}\subseteq\Simplex{\StateSpace\times\JointHistorySpace{t}\times\PublicHistorySpace{t}}\).
Although \(\UninfState{t}\) is not observed by either player, it is Markovian for a central planner.
Given \(\UninfState{t}\) and \(\JointPolicy{t}\), the next occupancy is \(\RedTrans{t}(\UninfState{t},\JointPolicy{t})=\Prob_{\UninfState{t},\JointPolicy{t}}(\BigState{t+1},\BigJointHistory{t+1},\BigPublicHistory{t+1})\).
Player \(i\)'s one-step expected reward is \(\RedRew{i,t}(\UninfState{t},\JointPolicy{t})=\Ex_{\UninfState{t},\JointPolicy{t}}[\RewardFunction{i}(\BigState{t},\BigJointAction{t})]\).
The central occupancy state restores Bellman's principle for the zero-sum value.
With \(\OptimalPrivateValueFunction{1}{\Horizon}(\cdot)\defeq 0\), the central dynamic program is
\[
\textstyle
\OptimalPrivateValueFunction{1}{t}(\UninfState{t})
=
NE_{(\Policy{1}{t}, \Policy{2}{t})} \{ \RedRew{1,t}(\UninfState{t},\JointPolicy{t})
+
\OptimalPrivateValueFunction{1}{t+1}(\RedTrans{t}(\UninfState{t},\JointPolicy{t})) \}
\]

where $NE$ denotes the operator that returns the value associated to the Nash equilibrium of the corresponding subgame.

This recursion is the basis of exact and approximate occupancy-based planning methods for Dec-POMDPs and zs-POSGs \citep{dibangoye2012scaling,DibangoyeABC13,dibangoye2014exploiting,Dibangoye2016,Dibangoye:OMDP:2016,pmlr-v80-dibangoye18a,DelBufDibSaf-DGAA-23,peralez2025optimally,peralezsolving,Escudie2025}.
The difficulty is not the recursion itself, but the space on which it acts.
The number of coordinates needed to represent \(\UninfState{t}\) grows exponentially with the horizon.
Point-based value iteration is a natural approximation paradigm for such continuous-state dynamic programs, but its zs-POSG instantiations provide limited guidance on where to place samples \citep{pineau2003point,SmithSimmons2004HSVI,ShaniPineauKaplow2013,DelBufDibSaf-DGAA-23,Escudie2025}.

The central planner game provides the correct value equation and, in principle, produces the optimal decision rules. The difficulty is instead computational: the central occupancy state induces a very large decision space. In particular, solving the corresponding subgame requires optimizing over decision rules defined on the possible central states, which can be viewed as a large matrix whose rows correspond to decision rules. This representation quickly becomes intractable as the horizon and information spaces grow.

The challenge lies in exploiting the decision rules produced by the central formulation without having to explicitly enumerate or sample the entire central occupancy space. In particular, we seek a representation that retains the correct Bellman recursion while allowing the relevant parts of the decision-rule space to be explored efficiently.

\section{Focal Belief Games}
\label{sec:focal:belief:games}

We rewrite the central state by conditioning on the information relevant to the focal player $\Player{i}$ and on the policies that determine the other players' behaviour. More precisely, we condition on the public history, the opponent's private history, and the focal player's policy prefix. This representation allows us to separate two steps of the solution procedure. First, during policy evaluation, we fix the focal player's current policy and compute the value it can guarantee against an opponent that best responds to it. This gives a pessimistic continuation value for the focal player's policy. Second, during policy improvement, this continuation value is used to determine a better decision rule for the focal player at the current stage, in the same spirit as a Bellman backup.


\begin{figure}[t]
    \centering
\begin{tikzpicture}[scale=1.1]

\draw[->, thick] (-0.2,1) -- (6,1) node[right] {$\BeliefSpace{i}{t}$};

\draw[gray!60]
    (0.5,1.5) -- (5.5,2.1);

\draw[gray!60]
    (0.5,3.2) -- (5.5,1.1);

\draw[gray!60]
    (0.5,2.9) -- (5.5,1.5);

\draw[dashed]
    (4,1) -- (4,2.);

\draw[gray!100, thick]
    (0.5,3.2)
    -- (2.6,2.32)
    -- (3.7,2)
    -- (4.4,1.97)
    -- (5.5,2.1);

\draw[thick]
    (0.5,2.2) -- (5.5,1.9);

\fill (4,1) circle (1.8pt)
    node[below right] {$\BeliefState{i}{t}$};

\node[align=center, anchor=west] at (5.5,2.) {$\underline{v}_{i,t}^{\pi^i_{t:\Horizon}}$};

\node at (1.5,3.2)
    {$\underline{v}_{i,t}$};


\end{tikzpicture}
\caption{The optimal value function of the game $\FocalBeliefGameTuple{i}$ is convex in the belief space. While value-iteration methods such as PBVI sample this space and operate on a set of vectors to represent $\underline{v}_{i,t}$, our method works with a single vector $\underline{v}_{i,t}^{\pi^i_{t:\Horizon}}$ at a time, reducing the complexity of the associated algorithms.}
\label{fig:pwlc_value}
\end{figure}
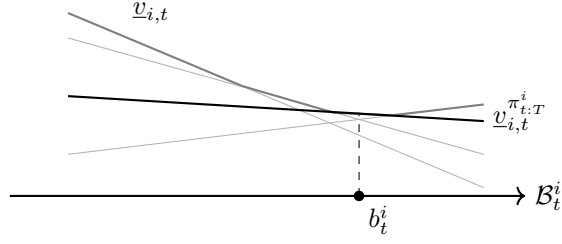

\subsection{Conditional occupancy states}

Fix focal player \(i\).
At stage \(t\), fix the public history \(\PublicHistory{t}\), the opponent private history \(\History{-i}{t}\), and the focal policy prefix \(\PrefixPolicy{i}{t}\).
Conditioning the central occupancy state on these variables gives \emph{the conditional occupancy state}
\[
\ConditionalState{i}{t}(\PublicHistory{t},\History{-i}{t},\PrefixPolicy{i}{t})
\defeq
\Prob_{\InitialBelief,\TransitionFunction}
\!\left(
\BigState{t},\BigHistory{i}{t}
\mid
\BigPublicHistory{t}=\PublicHistory{t},
\BigHistory{-i}{t}=\History{-i}{t},
\PrefixPolicy{i}{t}
\right).
\]
We write \(\ConditionalState{i}{t}\) when the histories and prefix are clear.
It is a distribution over the hidden state and the focal player's private history.
If the conditioning event has probability zero, the conditional update is fixed arbitrarily.

Let \(\ConditionalSpace{i}{t}\) be the set of reachable conditional occupancy states at stage \(t\).
In this conditional representation, the opponent's plan-time decision rule is written as 
\(\Policy{-i}{t}\colon\ConditionalSpace{i}{t}\to\ActionSpace{-i}{}\), with space \(\PolicySpace{-i}{t}\).
It does not denote an execution-time policy of the original zs-POSG. 
We use this representation for fixed-policy evaluation and improvement.
For fixed-policy evaluation, pure opponent actions are sufficient: after fixing the focal player's policy suffix, the objective is affine in the opponent's mixed action, i.e., in its distribution over $\ActionSpace{-i}{}$. Hence, the minimum over mixed actions is attained by a pure action.

Given \(\ConditionalState{i}{t}\), \(\Policy{i}{t}\), and \(\Action{-i}{t}\), the model induces a probability kernel
\(\TransitionCondKernel{i}{t}(\cdot\mid\ConditionalState{i}{t},\Policy{i}{t},\Action{-i}{t})\)
on \(\ConditionalSpace{i}{t+1}\).
This kernel is obtained by conditioning on the next public observation and the next opponent private observation.
The conditional Bayes update and its observation likelihood are denoted by \(\RetCondTrans{i}{t}\) and \(\RetCondObs{i}{t}\), and are given in Appendix~\ref{app:conditional-kernels}.
The one-step conditional payoff is
\(\RetCondRew{i}{t}(\ConditionalState{i}{t},\Policy{i}{t},\Action{-i}{t})\).

For a fixed focal policy suffix \(\SuffixPolicy{i}{t}\), these objects define the ingredients for a fixed-policy evaluation.
The policy suffix induces a decision rule \(\Policy{i}{k}\) at every stage \(k\ge t\).
With terminal value \(\PrivValFunc{i}{\Horizon}{\cdot}\equiv 0\), the fixed-policy value function satisfies 
\[
\PrivValFunc{i}{k}{\SuffixPolicy{i}{t}}(\ConditionalState{i}{k})
=
\min_{\Action{-i}{k}}
\textstyle
\left[
\RetCondRew{i}{k}(\ConditionalState{i}{k},\Policy{i}{k},\Action{-i}{k})
+
\sum_{\ConditionalState{i}{k+1}\in\ConditionalSpace{i}{k+1}}
\TransitionCondKernel{i}{k}(\ConditionalState{i}{k+1}\mid\ConditionalState{i}{k},\Policy{i}{k},\Action{-i}{k})
\PrivValFunc{i}{k+1}{\SuffixPolicy{i}{k+1}}(\ConditionalState{i}{k+1})
\right].
\]
This recursion evaluates a fixed focal policy suffix by accumulating one-step conditional payoffs.
It is the value function used later as the continuation term in local policy improvement.

\subsection{The focal belief game}

We now lift the conditional occupancy states to distributions over such states.
This gives the player-specific planning state used below.

Fix a focal player \(\Player{i}\).
The distribution over conditional occupancy state induced at stage \(t\) \emph{the focal belief}.
A focal belief \(\BeliefState{i}{t}\in\BeliefSpace{i}{t}\) is a probability distribution over \(\ConditionalSpace{i}{t}\).
Given \(\BeliefState{i}{t}\), the focal player chooses \(\Policy{i}{t}\).
The opponent's rule is represented in the conditional coordinates by \(\Policy{-i}{t}\).
The transition \(\RetBeliefTrans{i}{t}\) is the law of the next focal belief under \((\BeliefState{i}{t},\Policy{i}{t},\Policy{-i}{t})\).
The reward \(\RetBeliefRew{i}{t}\) is the expected conditional payoff under the same belief and decision rules.
Both are given in Appendix~\ref{app:belief-dynamics}


\begin{definition}
\(
\FocalBeliefGameTuple{i}
\doteq
(
\Agents,
(\BeliefSpace{i}{t})_{t=0}^{\Horizon},
(\PolicySpace{i}{t})_{t=0}^{\Horizon-1},
(\PolicySpace{-i}{t})_{t=0}^{\Horizon-1},
(\RetBeliefTrans{i}{t})_{t=0}^{\Horizon-1},
(\RetBeliefRew{i}{t})_{t=0}^{\Horizon-1},
\BeliefState{i}{0},
\Horizon
)
\)
is the focal belief game associated with the original game \(\GameTuple\).
\end{definition}

The focal belief game is fully observed at the planning level.
It is not a strict lossless reduction of the original zs-POSG  \(\GameTuple\).
The opponent's decision problem is represented in conditional coordinates and can therefore be viewed as a virtual player.
Fixed-policy value function have the value in both representations, and the focal decision-rule space is unchanged.
%
The problem is inherently one-sided: we fix the policy of a focal player and evaluate how well this player performs against an opponent that minimizes the remaining payoff. For a fixed focal policy suffix $\SuffixPolicy{i}{t}$, let $\ValueFunctionErwan{}_{i,t}^{\SuffixPolicy{i}{t}}(\BeliefState{i}{t})$ denote the value obtained from $\BeliefState{i}{t}$ when player $i$ follows $\SuffixPolicy{i}{t}$ and the opponent chooses actions to minimize the remaining payoff. This one-sided fixed-policy value function is the value function used by CPI.

\subsection{Structure of the focal belief value}

The focal belief game reveals structural properties of the value function that are not explicit in the set-valued focal state representation, as the theorem below states. In particular, the fixed-policy value function $\ValueFunctionErwan{}_{i,t}^{\SuffixPolicy{i}{t}}$ defines a linear function over the belief space $\BeliefSpace{i}{t}$.

\begin{theorem}
\label{thm:fixed_policy_linearity}
For any fixed focal policy suffix \(\SuffixPolicy{i}{t}\), the function
\(\ValueFunctionErwan{}_{i,t}^{\SuffixPolicy{i}{t}}\) is linear on \(\BeliefSpace{i}{t}\) and satisfies
\[
\textstyle
\ValueFunctionErwan{}_{i,t}^{\SuffixPolicy{i}{t}}(\BeliefState{i}{t})
=
\sum_{\ConditionalState{i}{t}\in \mathtt{supp}(\BeliefState{i}{t})}
\BeliefState{i}{t}(\ConditionalState{i}{t})
\PrivValFunc{i}{t}{\SuffixPolicy{i}{t}}(\ConditionalState{i}{t}).
\]
\end{theorem}

Although beliefs $\BeliefState{i}{t}$ typically have sparse support, their representation remains intractable due to the continuous nature of the conditional state space $\ConditionalSpace{i}{t}$. Nevertheless, this complexity can be mitigated: the value function induced by a policy $\SuffixPolicy{i}{t}$ can be exactly characterized using a reduced representation, relying only on a subset of relevant components.

\begin{proposition}
\label{prop:pwlc:structure}
For any fixed focal policy suffix $\SuffixPolicy{i}{t}$, 
$\PrivValFunc{i}{t}{\SuffixPolicy{i}{t}}$ is piecewise linear and concave on $\BeliefSpace{i}{t}$.
That is, there exists a finite collection $\Lambda^{\SuffixPolicy{i}{t}}_{i,t}$ of $\alpha$-vectors
$\alpha^{\SuffixPolicy{i}{t}}_{i,t} \in \mathbb{R}^{\BigState{t} \times \BigHistory{i}{t}}$,
each defining a linear function over the ConditionalSpace space $\ConditionalSpace{i}{t}$, such that:
\[
\textstyle
\PrivValFunc{i}{t}{\SuffixPolicy{i}{t}}(\ConditionalState{i}{t})
=
\min_{\alpha^{\SuffixPolicy{i}{t}}_{i,t} \in \Lambda^{\SuffixPolicy{i}{t}}_{i,t}}
\alpha^{\SuffixPolicy{i}{t}}_{i,t} \cdot \ConditionalState{i}{t}.
\]
\end{proposition}




\section{Conservative Policy Iteration}

We describe Conservative Policy Iteration (CPI) for finite-horizon zs-POSGs. CPI maintains one behavioural policy per player. At each iteration, these policies generate an occupancy trajectory after conservative averaging and GLIE exploration. For each player, it evaluates the current policy through the fixed-policy value function of Section~\ref{sec:focal:belief:games}. The resulting value function is used as the continuation term in a local Bellman backup. The backup returns an improved decision rule for the focal player. This rule is averaged into the current policy instead of being installed directly. CPI therefore performs local policy improvement along sampled occupancy trajectories. It does not construct the optimal value function of the focal belief game. It also does not attempt to cover the full central occupancy space.

\subsection{Local policy improvement}

Fix a stage \(t\), a focal belief \(\BeliefState{i}{t}\), and a current focal policy suffix \(\SuffixPolicy{i}{t+1}\). The fixed-policy value function \(\ValueFunctionErwan_{i,t+1}^{\SuffixPolicy{i}{t+1}}\) is used as the continuation value function after stage \(t\). We write \(\BeliefQVal{i}{t}\) for the corresponding local Q-value, defined by 
\(
\BeliefQVal{i}{t}(\BeliefState{i}{t}, \JointPolicy{t}) 
\defeq 
\RetBeliefRew{i}{t}(\BeliefState{i}{t}, \JointPolicy{t})
+
\ValueFunctionErwan_{i,t+1}^{\SuffixPolicy{i}{t+1}}(\RetBeliefTrans{i}{t}(\BeliefState{i}{t}, \JointPolicy{t}))
\).

The Bellman backup at \((t,\BeliefState{i}{t})\) induces a local zero-sum continuation game, \ie, the finite zero-sum game rooted at \((t,\BeliefState{i}{t})\) with a fixed-policy value function. The focal player chooses a decision rule \(\Policy{i}{t}\). The opponent chooses a decision rule \(\Policy{-i}{t}\) on conditional states. The payoff is \(\BeliefQVal{i}{t}(\BeliefState{i}{t},(\Policy{i}{t},\Policy{-i}{t}))\). We denote this local continuation game by \(\LocalGameTuple{i}{t}(\BeliefQVal{i}{t}(\BeliefState{i}{t},\cdot))\). We write \(\LocalGameTuple{i}{t}\) when the continuation value is clear from context.

\begin{definition}
Tuple \(\LocalGameTuple{i}{t} \doteq (\Agents, \PolicySpace{i}{t}, \PolicySpace{-i}{t},  \BeliefQVal{i}{t}(\BeliefState{i}{t},\cdot))\) is the finite zero-sum game with players \(\Agents\), action spaces \(\PolicySpace{i}{t}\) and \(\PolicySpace{-i}{t}\), and payoff \(\BeliefQVal{i}{t}(\BeliefState{i}{t},\cdot)\).
\end{definition}

Revisiting stage \(t\) at another belief therefore defines another local continuation game. It does not perturb the previous one. For fixed \(\BeliefQVal{i}{t}\), the local backup at \(\BeliefState{i}{t}\) is 
\(
\max_{\Policy{i}{t}}
\min_{\Policy{-i}{t}}
\BeliefQVal{i}{t}(\BeliefState{i}{t},(\Policy{i}{t},\Policy{-i}{t}))
\).
This is the local Bellman optimisation problem represented by \(\LocalGameTuple{i}{t}(\BeliefQVal{i}{t}(\BeliefState{i}{t},\cdot))\). A local solution is a max--min optimal focal decision rule together with an opponent minimising decision rule (see Theorem~\ref{thm:LP}). Since histories, observations, and actions are finite, deterministic local decision rules are finite. The focal planner may commit to mixed decision rules. The opponent admits deterministic minimising decision rules because the local payoff is affine in its mixed rule.

\subsection{Conservative Policy Iteration}

We now introduce conservative policy iteration (CPI). The algorithm alternes central occupancy expansion with local policy improvement in the two player-specific conditional representations.
\Cref{alg:api} maintains a fixed-policy value representation, step sizes, and averaged decision rules for both players; see Lines~\ref{alg:api:init-value}--\ref{alg:api:init-rules}.
Each iteration starts from the initial central occupancy state \(\UninfState{0}\) and calls \(\mathtt{Recurse}\); see Line~\ref{alg:api:call-recurse}.
The recursion propagates the central occupancy state because the joint policy acts on the original zs-POSG. Line~\ref{alg:api:expand}  expands then central occupancy state using the current joint decision rule.
At each reached \(\UninfState{t}\), Line~\ref{alg:api:belief} constructs, for each player \(i\), the focal belief induced by conditioning \(\UninfState{t}\) in player \(i\)'s basis
Line~\ref{alg:api:lp} applies the LP backup (see Theorem~\ref{thm:LP}) at this focal belief and returns an improved decision rule for each player.
Line~\ref{alg:api:average}--\ref{alg:api:glie} average the improved decision rule into the current decision rule and add GLIE exploration. GLIE is used to ensure that all components of the belief remain represented. Without exploration, the policy may assign zero probability to some actions, causing the induced belief to become degenerate and eliminating support from parts of the distribution. Maintaining exploration therefore prevents components of the belief from vanishing entirely, similarly to the distribution restart mechanism used in \cite{Kakade2002}. Line~\ref{alg:api:vectors-computed} computes the vector associated with the decision rule just computed, as established by Proposition~\ref{prop:pwlc:structure}.


\begin{algorithm}[t]
\caption{Conservative policy iteration.}
\label{alg:api}
\begin{minipage}[t]{0.45\linewidth}
\begin{algorithmic}[1]
\STATE Initialise \((\Lambda_{i,t})_{ t\in \{0,\ldots,\Horizon\},  i\in \Agents}\) \label{alg:api:init-value}
\STATE Initialise \((\alpha_k)_{k\ge0}\), \((\epsilon_k)_{k\ge0}\), \(k\gets0\) \label{alg:api:init-steps}
\STATE Initialise \((\AlgConDecRule{i}{t}{},u_{i,t})_{ t\in \{0,\ldots,\Horizon-1\}, i\in \Agents}\) \label{alg:api:init-rules}
\WHILE{not converged}
    \STATE \(\mathtt{Recurse}(\UninfState{0},k)\) \label{alg:api:call-recurse}
    \STATE \(\Policy{i}{t}{} \gets \AlgExpDecRule{i}{t}{}, \forall t \in \{0,...,\Horizon-1\}\) 
    \STATE \(k\gets k+1\)
\ENDWHILE
\STATE
\STATE \textbf{Function } \(\mathtt{CGLIE}(\Policy{i}{t},k)\colon\)
\STATE \quad \(\AlgConDecRule{i}{t}{} \gets(1-\alpha_k)\AlgConDecRule{i}{t}{} +\alpha_k \Policy{i}{t}\) \label{alg:api:average}
\STATE \quad \textbf{return} \((1-\epsilon_k)\AlgConDecRule{i}{t}{} +\epsilon_k u_{i,t}\) \label{alg:api:glie}
\end{algorithmic}
\end{minipage}
\hfill
\begin{minipage}[t]{0.58\linewidth}
\begin{algorithmic}[1]
\STATE \textbf{Function } \(\mathtt{Recurse}(\UninfState{t},k)\colon\)
\IF{\(t<\Horizon\)}
    \STATE \(\mathtt{Recurse}(\RedTrans{t}(\UninfState{t},(\Policy{1}{t}{}, \Policy{2}{t}{})),k)\) \label{alg:api:expand}
    \STATE \(\Policy{i}{t}{} \gets \mathtt{LP}(\mathtt{BELIEF}_i(\UninfState{t}),\Lambda_{i,t+1}), \forall i\in \Agents\). \label{alg:api:lp}
    \STATE \(\AlgExpDecRule{i}{t}{} \gets \mathtt{CGLIE}(\Policy{i}{t}{},k), \forall i\in \Agents\) \label{alg:api:cglie-call}
    \STATE \(\Lambda_{i,t} \gets \mathtt{VECTOR}(\AlgExpDecRule{i}{t}{}), \forall i\in \Agents\) \label{alg:api:vectors-computed}
\ENDIF
\STATE
\STATE \textbf{Function } \(\mathtt{BELIEF}_i(\UninfState{t})\colon\)
\STATE \quad \textbf{return} \(\Prob_{\UninfState{t}}(\BigConditionalState{i}{t})\) \label{alg:api:belief}
\end{algorithmic}
\end{minipage}
\end{algorithm}

\paragraph{Linear program.}
For a reached belief \(\BeliefState{i}{t}\), the local LP is written at the conditional level.
The input \(\Lambda_{i,t+1}\) denotes the stored lower envelop of \(\alpha\)-vectors representing 
the fixed-policy value function used for the continuation term. 
The LP implements the local backup in \(\LocalGameTuple{i}{t}\), \cf~Appendix~\ref{app:point-based-bellman-backups}.
In Algorithm~\ref{alg:api}, \(\mathtt{LP}(\BeliefState{i}{t}, \lambda_{i,t+1})\) denotes this local policy-improvement routine.

\begin{theorem}
\label{thm:algorithm_convergence}
Let the policy updates of player $i$ be given by
\[
    \Policy{i,k+1}{}
    =
    (1 - \alpha_k) \Policy{i,k}{}
    +
    \alpha_k
    \tilde{\Dynamics{i}{}}
    \left(
        \Policy{i,k}{},
        \Policy{-i,k}{}
    \right)
    +
    U_{i,k+1},
\]
where $\tilde{\Dynamics{i}{}}$ represents all the stage-wise policy optimization operator $\tilde{\Dynamics{i}{t}}$ obtained from the corresponding LP (Line~\ref{alg:api:lp} in Algorithm~\ref{alg:api}), and $U_{i,k+1}$ is a noise term. Assume that the step sizes satisfy the Robbins-Monro conditions
\[
    \sum_{k=0}^{\infty} \alpha_k = \infty,
    \qquad
    \sum_{k=0}^{\infty} \alpha_k^2 < \infty.
\]
Under the standard stochastic approximation conditions on $U_{i,k}$ and the approximation error between the discrete operator $\tilde{\Dynamics{i}{}}$ and the continuous dynamics $\Dynamics{i}{}$ (see Definition~\ref{def:continuous_dynamics}), the algorithm converges almost surely to the set of Nash equilibria.
\end{theorem}

Theorem~\ref{thm:algorithm_convergence} establishes the convergence of Algorithm~\ref{alg:api}, since its discrete update operator coincides with the operator defined in the algorithm. The GLIE noise introduced during exploration does not affect this convergence result, as it satisfies the conditions required by the convergence theorem.

\section{Empirical Evaluation}

We evaluate CPI on standard finite-horizon zero-sum POSG benchmarks: \textit{Adversarial Tiger}, \textit{Recycling}, \textit{Mabc}, and \textit{Matching Pennies}. Benchmark descriptions are available at \url{http://masplan.org/}. Some instances originate from common-payoff problems and are converted to the zero-sum setting by negating player~\(1\)'s payoff for player~\(2\). We compare CPI with PBVI \citep{Escudie2025}, HSVI \citep{DelBufDibSaf-DGAA-23} and CFR+\citep{Zinkevich2007,Tammelin14}.
The CFR+ baseline adapts the extensive-form game CFR+ directly to the occupancy-state reduction of zs-POSGs; implementation details are given in Appendix~\ref{appendix:ocfr}. In contrast, our CFR+ formulation naturally leverages the objects introduced in this paper, further strengthening the theoretical contribution of the paper. In particular, the POSG setting provides a compact representation of histories and occupancy states that can be exploited to speed up CFR. This allows CFR+ to reason directly over clusters of probabilistically equivalent private histories, rather than operating on the significantly larger extensive-form representation. All methods are evaluated on the same finite-horizon POSG instances. Table~\ref{table:main_results:bigtable} reports runtime, value, and exploitability. Results for extensive-form games are discussed solely in Appendix~\ref{extensive_form_game}.

\textbf{CPI vs PBVI.}
CPI is competitive with PBVI at small horizons and improves the runtime--exploitability trade-off at larger horizons. Indeed, PBVI grows exponentially with the horizon, as it adds an exponentially increasing number of occupancy states by construction. This difference is particularly visible in games such as Matching Pennies ($T=25,50$ in Table~\ref{table:main_results:bigtable} and Figure~\ref{fig:anytime_exp_value}), where PBVI is sensitive to the sampled belief set. For instance, insufficient or poorly distributed samples may slow convergence or degrade the quality of its central value function approximation. This sensitivity is particularly relevant because exploitability is approximated in our experiments: PBVI must converge to sufficiently low exploitability for this approximation to provide a reliable indication of safety, so a temporary increase in measured exploitability can indicate that the policy remains exploitable. In contrast, CPI updates policies along trajectories induced by the current conservative joint policy, which can guide updates toward regions where the policies are still exploitable. From a computational perspective, PBVI applies a relatively expensive backup operator over a growing collection of sampled belief points. As the horizon increases, maintaining and updating this point set becomes prohibitive. CPI instead performs local policy updates along sampled trajectories, avoiding the explicit construction of large point sets. In our experiments, this generally resulted in more stable scaling behaviour (\Cref{app:LP:PBVIvsAPI}).

Finally, although CPI and CFR+ outperform PBVI for higher horizon, PBVI achieves stronger results on Adversarial Tiger and Recycling (see Figure~\ref{fig:anytime_exp_value}). Adversarial Tiger is particularly well suited to methods that cover a broader occupancy space such as point-based value iteration algorithms. Notably, computation time increases proportionally with the horizon for CPI, whereas PBVI's computation time may vary depending on the sampling, as observed for Recycling in Table~\ref{table:main_results:bigtable}.

\newcommand{\subfigsize}{95pt}
\newcommand{\subfigfontsize}{\footnotesize}

\begin{figure*}
\centering


\begin{tabular}{c c c@{\hspace{-1pt}}c@{\hspace{-5pt}}c}
& \hspace{2em} \raisebox{0.5em}{\large\textbf{\textsc{\textcolor{gray}{Mabc}}}} & \hspace{2em} \raisebox{0.5em}{\large\textbf{\textsc{\textcolor{gray}{\shortstack{ Matching \\ Pennies}}}}} & \hspace{2em} \raisebox{0.5em}{\large\textbf{\textsc{\textcolor{gray}{\shortstack{Recycling}}}}} & \hspace{2em} \raisebox{0.5em}{\large\textbf{\textsc{\textcolor{gray}{\shortstack{Adversarial \\ Tiger}}}}}\\

\raisebox{1.em}{\rotatebox{90}{{\large\textbf{\textsc{\textcolor{gray}{Exploit.}}}}}} &
\begin{subfigure}{0.18\textwidth}
\centering
    \begin{tikzpicture}
      \begin{axis}[
        title style={yshift=-20pt, fill=white},
        width=\subfigsize,
        height=\subfigsize,
        xmin=4, xmax=100,
        xtick={0, 100},
        ymin=0, ymax=0.2,
        ytick={0, 0.1, 0.2},
        ylabel style={yshift=-6pt,font=\subfigfontsize},
        xlabel style={yshift=4pt,font=\subfigfontsize},
        grid=both,
        scaled y ticks=false,
        yticklabel style={
          /pgf/number format/fixed,
          /pgf/number format/fixed zerofill,
          /pgf/number format/precision=2,
         font=\subfigfontsize
        },
        style={very thick,font=\subfigfontsize},
        axis line style={very thick},
        legend style={draw=none}
      ]
        \addplot+[const plot, mark=none, very thick, sthlmGreen, opacity=1., solid] table [col sep=comma, x=cumul_time, y=exp_value] {./paper/results/seq_mabc_H5_T2_S42.log};
        \addplot+[mark=none, very thick, sthlmRed, opacity=1., solid] table [col sep=comma, x=cumul_time, y=norm_exp_value] {./paper/results/cfr_mabc_H5_T2_S42.log};
        \addplot+[mark=none, very thick, sthlmYellow, opacity=1., solid] table [col sep=comma, x=cumul_time, y expr= -\thisrow{value} - \thisrow{S_per_step}] {./paper/results/api_mabc_H5_T1_S42_N5.log};
      \end{axis}
    \end{tikzpicture}
\end{subfigure} &
\begin{subfigure}{0.18\textwidth}
\centering
    \begin{tikzpicture}
      \begin{axis}[
        title style={yshift=-20pt, fill=white},
        width=\subfigsize,
        height=\subfigsize,
        xmin=0, xmax=30,
        xtick={0, 30},
        ymin=0, ymax=0.05,
        ytick={0, 0.025, 0.05},
        ylabel style={yshift=-6pt,font=\subfigfontsize},
        xlabel style={yshift=4pt,font=\subfigfontsize},
        grid=both,
        scaled y ticks=false,
        yticklabel style={
          /pgf/number format/fixed,
          /pgf/number format/fixed zerofill,
          /pgf/number format/precision=2,
         font=\subfigfontsize
        },
        style={very thick,font=\subfigfontsize},
        axis line style={very thick},
        legend style={draw=none}
      ]
        \addplot+[const plot, mark=none, very thick, sthlmGreen, opacity=1., solid] table [col sep=comma, x=cumul_time, y=exp_value] {./paper/results/seq_matchingpennies_H5_T1_S42.log};
        \addplot+[mark=none, very thick, sthlmYellow, opacity=1., solid] table [col sep=comma, x=cumul_time, y=exp_val] {./paper/results/api_matchingpennies_H5_T1_S42_N2.log};
      \end{axis}
    \end{tikzpicture}
\end{subfigure} &
\begin{subfigure}{0.18\textwidth}
    \centering
    \begin{tikzpicture}
      \begin{axis}[
        title style={yshift=-20pt, fill=white},
        width=\subfigsize,
        height=\subfigsize,
        xmin=0, xmax=400,
        xtick={0, 400},
        ymin=0, ymax=0.1,
        ytick={0, 0.05, 0.1},
        ylabel style={yshift=-6pt,font=\subfigfontsize},
        xlabel style={yshift=4pt,font=\subfigfontsize},
        grid=both,
        scaled y ticks=false,
        yticklabel style={
          /pgf/number format/fixed,
          /pgf/number format/fixed zerofill,
          /pgf/number format/precision=2,
         font=\subfigfontsize
        },
        style={very thick,font=\subfigfontsize},
        axis line style={very thick},
        legend style={draw=none}
      ]
        \addplot+[const plot, mark=none, very thick, sthlmGreen, opacity=1., solid] table [col sep=comma, x=cumul_time, y=exp_value] {./paper/results/seq_recycling_H5_T1_S42.log};
        \addplot+[mark=none, very thick, sthlmRed, opacity=1., solid] table [col sep=comma, x=cumul_time, y expr= 2*\thisrow{norm_exp_value}] {./paper/results/cfr_recycling_H5_T1_S42.log};
        \addplot+[mark=none, very thick, sthlmYellow, opacity=1., solid] table [col sep=comma, x=cumul_time, y expr= -\thisrow{value} - \thisrow{S_per_step}] {./paper/results/api_recycling_H5_T1_S42_N5.log};
      \end{axis}
    \end{tikzpicture}
\end{subfigure} &
\begin{subfigure}{0.18\textwidth}
\centering
    \begin{tikzpicture}
      \begin{axis}[
        title style={yshift=-20pt, fill=white},
        width=\subfigsize,
        height=\subfigsize,
        xmin=0, xmax=250,
        xtick={0, 250},
        ymin=0, ymax=0.15,
        ytick={0, 0.075, 0.15},
        ylabel style={yshift=-6pt,font=\subfigfontsize},
        xlabel style={yshift=4pt,font=\subfigfontsize},
        grid=both,
        scaled y ticks=false,
        yticklabel style={
          /pgf/number format/fixed,
          /pgf/number format/fixed zerofill,
          /pgf/number format/precision=2,
         font=\subfigfontsize
        },
        style={very thick,font=\subfigfontsize},
        axis line style={very thick},
        legend style={draw=none}
      ]
        \addplot+[const plot, mark=none, very thick, sthlmGreen, opacity=1., solid] table [col sep=comma, x=cumul_time, y=exp_value] {./paper/results/seq_adversarialtiger_H5_T1_S42.log};
        \addplot+[mark=none, very thick, sthlmRed, opacity=1., solid] table [col sep=comma, x=cumul_time,y expr= 2*\thisrow{norm_exp_value}] {./paper/results/cfr_adversarialtiger_H5_T1_S42.log};
        \addplot+[mark=none, very thick, sthlmYellow, opacity=1., solid] table [col sep=comma, x=cumul_time, y expr= -\thisrow{value} - \thisrow{S_per_step}] {./paper/results/api_adversarialtiger_H5_T1_S42_N10.log};
      \end{axis}
    \end{tikzpicture}
\end{subfigure}

\\[1em]

\raisebox{2.em}{\rotatebox{90}{{\large\textbf{\textsc{\textcolor{gray}{Value}}}}}} &
\begin{subfigure}{0.18\textwidth}
\centering
    \begin{tikzpicture}
      \begin{axis}[
        title style={yshift=-20pt, fill=white},
        width=\subfigsize,
        height=\subfigsize,
        xmin=4, xmax=100,
        xtick={0, 100},
        ymin=0, ymax=0.2,
        ytick={0, 0.1, 0.2},
        ylabel style={yshift=-6pt,font=\subfigfontsize},
        xlabel style={yshift=4pt,font=\subfigfontsize},
        grid=both,
        scaled y ticks=false,
        yticklabel style={
          /pgf/number format/fixed,
          /pgf/number format/fixed zerofill,
          /pgf/number format/precision=2,
         font=\subfigfontsize
        },
        style={very thick,font=\subfigfontsize},
        axis line style={very thick},
        legend style={draw=none}
      ]
        \addplot+[const plot, mark=none, very thick, sthlmGreen, opacity=1., solid] table [col sep=comma, x=cumul_time, y=value] {./paper/results/seq_mabc_H5_T2_S42.log};
        \addplot+[mark=none, very thick, sthlmRed, opacity=1., solid] table [col sep=comma, x=cumul_time, y=value] {./paper/results/cfr_mabc_H5_T2_S42.log};
        \addplot+[mark=none, very thick, sthlmYellow, opacity=1., solid] table [col sep=comma, x=cumul_time, y=value] {./paper/results/api_mabc_H5_T2_S42_N2.log};
      \end{axis}
    \end{tikzpicture}
\end{subfigure} &
\begin{subfigure}{0.18\textwidth}
\centering
    \begin{tikzpicture}
      \begin{axis}[
        title style={yshift=-20pt, fill=white},
        width=\subfigsize,
        height=\subfigsize,
        xmin=0, xmax=30,
        xtick={0, 30},
        ymin=0.5, ymax=1.5,
        ytick={0.5, 1., 1.5},
        ylabel style={yshift=-6pt,font=\subfigfontsize},
        xlabel style={yshift=4pt,font=\subfigfontsize},
        grid=both,
        scaled y ticks=false,
        yticklabel style={
          /pgf/number format/fixed,
          /pgf/number format/fixed zerofill,
          /pgf/number format/precision=2,
         font=\subfigfontsize
        },
        style={very thick,font=\subfigfontsize},
        axis line style={very thick},
        legend style={draw=none}
      ]
        \addplot+[const plot, mark=none, very thick, sthlmGreen, opacity=1., solid] table [col sep=comma, x=cumul_time, y=value] {./paper/results/seq_matchingpennies_H5_T1_S42.log};
        \addplot+[mark=none, very thick, sthlmRed, opacity=1., solid] table [col sep=comma, x=cumul_time, y=value] {./paper/results/cfr_matchingpennies_H5_T1_S42.log};
        \addplot+[mark=none, very thick, sthlmYellow, opacity=1., solid] table [col sep=comma, x=cumul_time, y=value] {./paper/results/api_matchingpennies_H5_T1_S42_N2.log};
      \end{axis}
    \end{tikzpicture}
\end{subfigure} &
\begin{subfigure}{0.18\textwidth}
\centering
    \begin{tikzpicture}
      \begin{axis}[
        title style={yshift=-20pt, fill=white},
        width=\subfigsize,
        height=\subfigsize,
        xmin=0, xmax=400,
        xtick={0, 400},
        ymin=0.3, ymax=0.5,
        ytick={0.3, 0.4, 0.5},
        ylabel style={yshift=-6pt,font=\subfigfontsize},
        xlabel style={yshift=4pt,font=\subfigfontsize},
        grid=both,
        scaled y ticks=false,
        yticklabel style={
          /pgf/number format/fixed,
          /pgf/number format/fixed zerofill,
          /pgf/number format/precision=2,
         font=\subfigfontsize
        },
        style={very thick,font=\subfigfontsize},
        axis line style={very thick},
        legend style={draw=none}
      ]
        \addplot+[const plot, mark=none, very thick, sthlmGreen, opacity=1., solid] table [col sep=comma, x=cumul_time, y=value] {./paper/results/seq_recycling_H5_T1_S42.log};
        \addplot+[mark=none, very thick, sthlmRed, opacity=1., solid] table [col sep=comma, x=cumul_time, y=value] {./paper/results/cfr_recycling_H5_T1_S42.log};
        \addplot+[mark=none, very thick, sthlmYellow, opacity=1., solid] table [col sep=comma, x=cumul_time, y=value] {./paper/results/api_recycling_H5_T1_S42_N2.log};
      \end{axis}
    \end{tikzpicture}
\end{subfigure} &
\begin{subfigure}{0.18\textwidth}
\centering
    \begin{tikzpicture}
      \begin{axis}[
        title style={yshift=-20pt, fill=white},
        width=\subfigsize,
        height=\subfigsize,
        xmin=0, xmax=250,
        xtick={0, 250},
        ymin=-1., ymax=-0.8,
        ytick={-1, -0.9, -0.8},
        ylabel style={yshift=-6pt,font=\subfigfontsize},
        xlabel style={yshift=4pt,font=\subfigfontsize},
        grid=both,
        scaled y ticks=false,
        yticklabel style={
          /pgf/number format/fixed,
          /pgf/number format/fixed zerofill,
          /pgf/number format/precision=1,
         font=\subfigfontsize
        },
        style={very thick,font=\subfigfontsize},
        axis line style={very thick},
        legend style={draw=none}
      ]
        \addplot+[const plot, mark=none, very thick, sthlmGreen, opacity=1., solid] table [col sep=comma, x=cumul_time, y=value] {./paper/results/seq_adversarialtiger_H5_T1_S42.log};
        \addplot+[mark=none, very thick, sthlmRed, opacity=1., solid] table [col sep=comma, x=cumul_time, y=value] {./paper/results/cfr_adversarialtiger_H5_T1_S42.log};
        \addplot+[mark=none, very thick, sthlmYellow, opacity=1., solid] table [col sep=comma, x=cumul_time, y=value] {./paper/results/api_adversarialtiger_H5_T1_S42_N1.log};
      \end{axis}
    \end{tikzpicture}
\end{subfigure}

\end{tabular}

\caption{Evolution of value and exploitability across different games under a fixed horizon $\ell = 5$. The results compare three methods: \textcolor{sthlmRed}{CFR+}, \textcolor{sthlmGreen}{PBVI}, and \textcolor{sthlmYellow}{CPI}, highlighting their respective performance profiles in terms of convergence behaviour and solution quality across the considered benchmarks.}
\label{fig:anytime_exp_value}
\end{figure*}
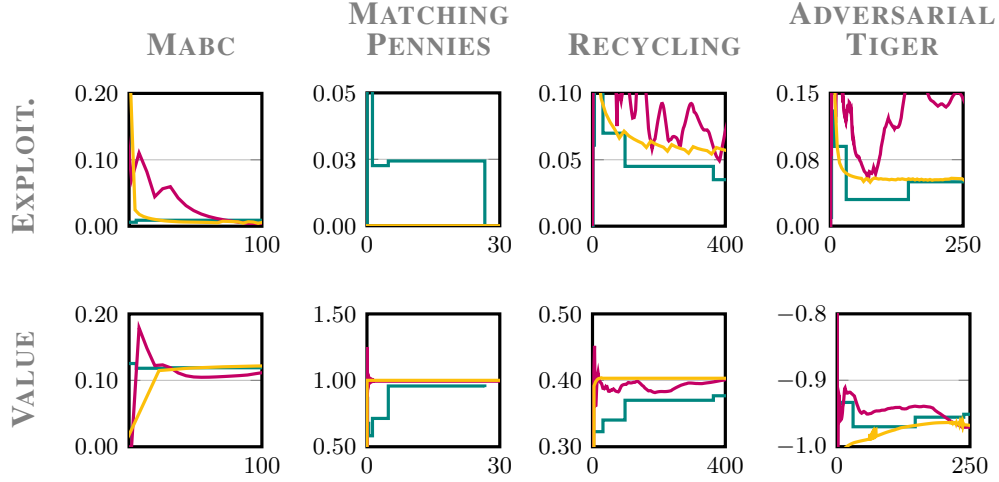
\definecolor{verylightgray}{gray}{0.9}

\begin{table}
    \setlength\extrarowheight{-3pt}

    \caption{Snapshot of the empirical results, including the computational cost of exploitability evaluation. For each game, horizon and algorithm, we report time, value and exploitability $\epsilon$ associated with the computed $\epsilon$-Nash equilibrium. {\sc OOT} occurs when the target exploitability threshold cannot be reached within the prescribed time limit, while {\sc oom} means that the algorithms ran out of memory.  \textcolor{magenta}{\textbf{magenta}} = lowest exploitability/value per benchmark}
    \label{table:main_results:bigtable}
    \resizebox{\linewidth}{!}{
        \begin{tabular}{@{}c c rr@{\hspace{4pt}}l rr@{\hspace{4pt}}l rr@{\hspace{4pt}}l rr@{\hspace{4pt}}l}
        
        \toprule%
         \centering%
         & 
         & \multicolumn{3}{c}{{{\bfseries \texttt{CPI} (Ours)}}}
         & \multicolumn{3}{c}{{{\bfseries \texttt{PBVI}}}}
         & \multicolumn{3}{c}{{{\bfseries \texttt{HSVI}}}}
         & \multicolumn{3}{c}{{{\bfseries \texttt{CFR+}}}}
         \\

         \multirow{-4}{*}{\textbf{Problem}} & \multirow{-4}{*}{$T$} & t & val. & $\epsilon$ & t & val. & $\epsilon$ & t & val. & $\epsilon$ & t & val. & $\epsilon$\\ 
        
        \cmidrule[0.4pt](lr{1em}){3-5}%
        \cmidrule[0.4pt](lr{1em}){6-8}%
        \cmidrule[0.4pt](lr{1em}){9-11}%
        \cmidrule[0.4pt](lr{1em}){12-14}%

        \arrayrulecolor{verylightgray}
        
            & 2 & {0.3} & \opti{0.078} & \opti{0.} & \opti{0.2} & \opti{0.078} & \opti{0.} & {0.4} & \opti{0.078} & \opti{0.} & {0.3} & \opti{0.078} & \opti{0.}\\

            
            & 3 & \opti{11} & \opti{0.097} & \opti{0.} & {24} & {0.096} & \opti{0.} & {70} & \opti{0.097} & \opti{0.} & {12} & {0.095} & {0.}\\
            
            
            & 4 & \opti{23} & \opti{0.11} & \opti{0.} & {65} & \opti{0.11} & {0.01} & \textsc{oot} & \textsc{oot} & \textsc{oot} & {49} & \opti{0.11} & {0.01}\\
            
            
            & 5 & \opti{29} & \opti{0.12} & \opti{0.} & {31} & \opti{0.12} & \opti{0.} & \textsc{oot} & \textsc{oot} & \textsc{oot} & {82} & \opti{0.12} & {0.01}\\
            
            
            & 7 & \opti{97} & \opti{0.14} & {0.01} & {190} & \opti{0.14} & \opti{0.005} & \textsc{oot} & \textsc{oot} & \textsc{oot} & {292} & \opti{0.14} & {0.02}\\
            
            
            & 10 & \opti{100} & \opti{0.18} & \opti{0.01} & {637} & {0.17} & \opti{0.01} & \textsc{oot} & \textsc{oot} & \textsc{oot} & {380} & \opti{0.18} & {0.02}\\
            
            
            
            
            \multirow{-11}{*}{\textsc{Mabc}} & 100 & \opti{1920} & \opti{1.06} & \opti{0.03} & \textsc{oot} & \textsc{oot} & \textsc{oot} & \textsc{oot} & \textsc{oot} & \textsc{oot} & {10042} & \opti{1.06} & \opti{0.2}\\
        
        \noalign{\vskip 2pt}
        \arrayrulecolor{black}
        \hline
        \noalign{\vskip 4pt}

            & 2 & \opti{0.01} & \opti{0.4} & \opti{0.} & {2} & {0.39} & \opti{0.} & {0.1} & \opti{0.4} & \opti{0.} & {0.06} & \opti{0.4} & {0.005}\\
            
            & 3 & \opti{0.02} & \opti{0.6} & \opti{0.} & {18} & \opti{0.6} & \opti{0.} & {1} & \opti{0.6} & \opti{0.} & {0.1} & \opti{0.6} & {0.001}\\
            
            & 4 & \opti{0.04} & \opti{0.8} & \opti{0.} & {34} & {0.77} & {0.} & {4} & \opti{0.8} & \opti{0.} & {0.2} & \opti{0.8} & \opti{0.}\\
            
            & 5 & \opti{0.04} & \opti{1.} & \opti{0.} & {52} & {0.95} & \opti{0.} & {60} & \opti{1.} & \opti{0.} & {0.25} & \opti{1.} & \opti{0.}\\
            
            & 7 & \opti{13} & \opti{1.4} & \opti{0.} & {180} & {1.31} & {0.03} & \textsc{oot} & \textsc{oot} & \textsc{oot} & {15} & {1.39} & {0.01}\\
            
            & 10 & \opti{92} & \opti{2.} & \opti{0.} & {\textsc{oot}} & \textsc{oot} & \textsc{oot} & \textsc{oot} & \textsc{oot} & \textsc{oot} & {103} & {1.99} & {0.02}\\

            & 25 & \opti{15} & \opti{5.} & \opti{0.} & \textsc{oot} & \textsc{oot} & \textsc{oot} & \textsc{oot} & \textsc{oot} & \textsc{oot} & {132} & \opti{5.} & {0.02}\\
            
            \multirow{-9}{*}{\shortstack[C]{\textsc{Matching} \\ \textsc{Pennies}}} & 50 & \opti{217} & \opti{10.} & \opti{0.} & \textsc{oot} & \textsc{oot} & \textsc{oot} & \textsc{oot} & \textsc{oot} & \textsc{oot} & {718} & {10.} & {0.03}\\

        \noalign{\vskip 2pt}
        \arrayrulecolor{black}
        \hline
        \noalign{\vskip 4pt}

            & 2 & \opti{0.4} & \opti{0.26} & \opti{0.} & \opti{0.4} & \opti{0.26} & \opti{0.} & {1.2} & \opti{0.26} & \opti{0.} & {2.8} & \opti{0.26} & {0.004}\\
            
            & 3 & \opti{17} & \opti{0.32} & {0.01} & {78} & \opti{0.32} & \opti{0.} & {120} & \opti{0.32} & \opti{0.} & {75} & \opti{0.32} & {0.01}\\
            
            & 4 & \opti{42} & \opti{0.36} & {0.03} & {56} & {0.35} & \opti{0.01} & \textsc{oot} & \textsc{oot} & \textsc{oot} & {139} & {0.35} & {0.04}\\
            
            & 5 & {213} & \opti{0.4} & \opti{0.05} & \opti{98} & {0.38} & \opti{0.05} & \textsc{oot} & \textsc{oot} & \textsc{oot} & {239} & {0.38} & {0.06}\\
            
            & 7 & \opti{271} & \opti{0.5} & {0.06} & {1481} & {0.48} & \opti{0.04} & \textsc{oot} & \textsc{oot} & \textsc{oot} & {624} & \opti{0.46} & {0.12}\\

            & 10 & \opti{505} & \opti{0.63} & {0.07} & {3232} & {0.6} & \opti{0.06} & \textsc{oot} & \textsc{oot} & \textsc{oot} & {1249} & \opti{0.63} & {0.15}\\
            
            \multirow{-9}{*}{\textsc{Recycling}} & {100} & \opti{4180} & {4.9} & \opti{0.68} & \textsc{oot} & \textsc{oot} & \textsc{oot} & \textsc{oot} & \textsc{oot} & \textsc{oot} & {13087} & \opti{4.96} & {2}\\

        \noalign{\vskip 2pt}
        \arrayrulecolor{black}
        \hline
        \noalign{\vskip 4pt}
        
            & 2 & \opti{0.1} & \opti{-0.4} & \opti{0.} & \opti{0.1} & \opti{-0.4} & \opti{0.} & {0.2} & \opti{-0.4} & \opti{0.} & \opti{0.1} & \opti{-0.4} & \opti{0.0}\\
            
            & 3 & \opti{0.2} & \opti{-0.56} & \opti{0.} & {0.3} & \opti{-0.56} & \opti{0.} & {500} & \opti{-0.56} & \opti{0.} & {5} & \opti{-0.56} & \opti{0.}\\
            
            & 4 & \opti{16} & \opti{-0.75} & \opti{0.01} & {46} & {-0.76} & {0.02} & \textsc{oot} & \textsc{oot} & \textsc{oot} & {76} & \opti{-0.75} & {0.03}\\

            & 5 & \opti{82} & {-0.96} & \opti{0.05} & {141} & {-0.96} & \opti{0.05} & \textsc{oot} & \textsc{oot} & \textsc{oot} & {91} & \opti{-0.95} & \opti{0.05}\\

            & 7 & \opti{248} & {-1.36} & {0.13} & {866} & {-1.36} & \opti{0.1} & \textsc{oot} & \textsc{oot} & \textsc{oot} & {459} & \opti{-1.35} & {0.13}\\
            
            \multirow{-7}{*}{\shortstack[C]{\textsc{Adversarial} \\ \textsc{Tiger}}} & 10 & \opti{824} & {-1.97} & {0.24} & {956} & {-1.98} & \opti{0.18} & \textsc{oot} & \textsc{oot} & \textsc{oot} & {3901} & \opti{-1.95} & {0.3}\\

        \bottomrule
        \end{tabular}
        \vspace{-.5cm}
    }
\end{table}

\textbf{CPI vs CFR.}
Both CPI and CFR-based methods improve policies iteratively, but through different update mechanisms.
In our experiments, CFR iterations are computationally cheaper than CPI updates. 
However, exploitability often evolves non-monotonically across iterations (see Figure~\ref{fig:anytime_exp_value} for Recycling and Adversarial Tiger).
This finding is consistent with observations reported in \citep{DelBufDibSaf-DGAA-23}. 
CPI instead performs more expensive local updates based on explicit continuation evaluations.
This generally produces smoother exploitability profiles in the reported experiments (see Figure~\ref{fig:anytime_exp_value}).
A practical difference concerns exploitability evaluation. 
For CFR, exploitability must be estimated through an additional evaluation procedure whose cost increases with horizon. For fairness, these evaluation costs are included in the reported runtimes (see Table~\ref{table:main_results:bigtable}). Runtimes excluding the measurement cost are reported in Table~\ref{table:main_results:bigtable:exploitability_cost}. In CPI, the quantities required to estimate deviations are maintained during policy improvement steps via the fixed-policy value function. So exploitability estimates are obtained directly from the algorithm.
For larger horizons, exploitability estimates in CFR also become increasingly noisy, which can make convergence trends more difficult to interpret empirically.
CFR+ remains attractive because of its inexpensive iterations, whereas CPI performs fewer but more costly updates.
The two approaches therefore reflect different computational trade-offs between update cost and exploitability estimation.

\paragraph{Takehome message.} 
 CFR+ benefits from inexpensive iterations, but typically requires more iterations and repeated exploitability evaluations to reach  solution quality comparable to CPI. 
 In our experiments, this often resulted in higher overall runtime at larger horizons.
 CPI instead performs fewer but more expensive updates, leading to a different trade-off between update cost, exploitability estimation, and empirical stability.















\section{Conclusion}

We introduced an approximate policy-iteration algorithm for finite-horizon zs-POSGs.
The method combines elements from point-based value iteration and policy-search approaches.
It maintains one policy per player and updates both policies along occupancy trajectories induced by conservative policy updates.
Instead of approximating a global central value function, each update relies on a fixed-policy continuation value function associated with the current joint policy.
The algorithm combines local policy improvements, conservative updates, and persistent exploration.
The empirical results show that this approach is competitive on standard finite-horizon zs-POSG benchmarks.
In several instances and horizons, it improves the runtime--exploitability trade-off compared with PBVI, HSVI, and CFR+.

Several limitations remain.
First, the algorithm requires solving local optimisation problems for both players throughout training.
Second, convergence can remain slow and the sample complexity high, which may limit scalability at larger horizons.
Third, the current implementation still relies on linear programming subroutines, which may become computationally demanding in larger games.
Future work will investigate iterative approximations of these optimisation steps, improved exploration and scheduling mechanisms, and neural approximations of the continuation value functions.

\subsection*{AI use statement}

In this work, we used generative AI tools to assist with translation, provide critical ingredients for proving mathematical claims, assist in the development and refinement of mathematical proofs, edit the manuscript to improve readability, create and modify scientific figures, create and edit software code, summarize and analyze existing literature, and assist with sourcing and searching for relevant information.

We did not use generative AI tools to develop theoretical models or conceptual frameworks, formulate mathematical claims, propose or refine hypotheses, design research methodology or experiments, provide feedback on research methodology or experiments, implement the methods underlying the research, or interpret the results. The remaining categories of required disclosure are not applicable to this work.

All AI-assisted content and outputs were reviewed by the authors. The authors take full responsibility for the final content of this work, including all text, mathematical claims, proofs, figures, code, and other artifacts produced with the assistance of generative AI.



\bibliography{iclr2027_conference}
\bibliographystyle{iclr2027_conference}



\newpage
\appendix

\section{Proofs}
\label{app:proofs}

\begin{lemma}
\label{lem:finite-compact-attainment}
In a finite-horizon finite model, all behavioural-policy spaces are compact, and all payoff maps are continuous.
\end{lemma}

\begin{proof}
Each \(\PolicySpace{i}{t}\) is a finite product of finite-dimensional simplexes.
Each \(\PolicySpace{i}{}\) and each \(\SuffixPolicySpace{i}{t}\) is a finite product of decision-rule spaces.
Finite products of compact sets are compact.
Every finite-horizon payoff is a finite sum of multilinear functions of behavioural-policy coordinates.
Hence every finite-horizon payoff map is continuous.
The extreme-value theorem gives attainment for every optimisation over these spaces.
\end{proof}

\label{app:proof-belief-reduction}

\subsection{Conditional kernels}
\label{app:conditional-kernels}

If a conditioning event has zero probability, \(\RetCondTrans{i}{t}\colon \ConditionalSpace{i}{t} \times \PolicySpace{i}{t} \times \ActionSpace{-i}{} \times \PublicObservationSpace{} \times \ObservationSpace{-i}{} \to \ConditionalSpace{i}{t+1}\) is assigned an arbitrary fixed element of \(\ConditionalSpace{i}{t+1}\).
All such terms are multiplied by their zero likelihood.

\begin{lemma}
\label{lem:cond-trans-bayes}
\(\RetCondTrans{i}{t}\) is the Bayes update of \((\BigState{t+1},\BigHistory{i}{t+1})\) given \((\BigPublicObservation{t+1},\BigObservation{-i}{t+1})\).
\end{lemma}

\begin{proof}
Fix \((\ConditionalState{i}{t},\Policy{i}{t},\Action{-i}{t},\PublicObservation{t+1},\Observation{-i}{t+1})\).

For \((\State{t+1},\History{i}{t+1})\), define
\[
\begin{aligned}
m(\State{t+1},\History{i}{t+1})
&\defeq
\textstyle
\sum_{\State{t},\History{i}{t},\Action{i}{t},\Observation{i}{t+1}}
\ConditionalState{i}{t}(\State{t},\History{i}{t})
\Policy{i}{t}(\Action{i}{t}\mid\History{i}{t})
\\
&\quad\cdot
\TransitionFunction(
\State{t+1},
\PublicObservation{t+1},
\Observation{i}{t+1},
\Observation{-i}{t+1}
\mid
\State{t},
\Action{i}{t},
\Action{-i}{t})
\cdot
\mathbf{1}\{
\History{i}{t+1}=(\History{i}{t},\Action{i}{t},\Observation{i}{t+1})
\}.
\end{aligned}
\]
Let \(W\defeq\sum_{\State{t+1},\History{i}{t+1}}m(\State{t+1},\History{i}{t+1})\).

If \(W>0\), then
\[
\RetCondTrans{i}{t}(\ConditionalState{i}{t},\Policy{i}{t},\Action{-i}{t},\PublicObservation{t+1},\Observation{-i}{t+1})(\State{t+1},\History{i}{t+1})
=
\frac{m(\State{t+1},\History{i}{t+1})}{W}.
\]
If \(W=0\), every weighted term involving this update is zero.
\end{proof}

\begin{lemma}
\label{lem:cond-obs-law}
\(\RetCondObs{i}{t} \colon \ConditionalSpace{i}{t} \times \PolicySpace{i}{t} \times \ActionSpace{-i}{}  \to \Simplex{\PublicObservationSpace{}\times\ObservationSpace{-i}{}}\) is a probability law over \(\PublicObservationSpace{}\times\ObservationSpace{-i}{}\).
\end{lemma}

\begin{proof}
Fix \((\ConditionalState{i}{t},\Policy{i}{t},\Action{-i}{t})\).

With \(m\) as in Lemma~\ref{lem:cond-trans-bayes}, define
\[
\RetCondObs{i}{t}(\ConditionalState{i}{t},\Policy{i}{t},\Action{-i}{t},\PublicObservation{t+1},\Observation{-i}{t+1})
\defeq
\sum_{\State{t+1},\History{i}{t+1}}
m(\State{t+1},\History{i}{t+1}).
\]
Then \(\RetCondObs{i}{t}(\ConditionalState{i}{t},\Policy{i}{t},\Action{-i}{t},\PublicObservation{t+1},\Observation{-i}{t+1})\ge0\), and
\[
\begin{aligned}
&\sum_{\PublicObservation{t+1},\Observation{-i}{t+1}}
\RetCondObs{i}{t}(\ConditionalState{i}{t},\Policy{i}{t},\Action{-i}{t},\PublicObservation{t+1},\Observation{-i}{t+1})
\\
&=
\sum_{\State{t},\History{i}{t}}
\ConditionalState{i}{t}(\State{t},\History{i}{t})
\sum_{\Action{i}{t}}
\Policy{i}{t}(\Action{i}{t}\mid\History{i}{t})
\cdot
\sum_{\State{t+1},\PublicObservation{t+1},\Observation{i}{t+1},\Observation{-i}{t+1}}
\TransitionFunction(
\State{t+1},
\PublicObservation{t+1},
\Observation{i}{t+1},
\Observation{-i}{t+1}
\mid
\State{t},
\Action{i}{t},
\Action{-i}{t})
\\
&=
\sum_{\State{t},\History{i}{t}}
\ConditionalState{i}{t}(\State{t},\History{i}{t})
\sum_{\Action{i}{t}}
\Policy{i}{t}(\Action{i}{t}\mid\History{i}{t})
=
1.
\end{aligned}
\]
\end{proof}

\begin{lemma}
\label{lem:cond-reward}
\(\RetCondRew{i}{t}\colon \ConditionalSpace{i}{t} \times \PolicySpace{i}{t} \times \ActionSpace{-i}{}  \to \mathbb{R}\) is the conditional expectation of \(\RewardFunction{i}(\BigState{t},\BigAction{i}{t},\Action{-i}{t})\).
\end{lemma}

\begin{proof}
Fix \((\ConditionalState{i}{t},\Policy{i}{t},\Action{-i}{t})\).

\[
\begin{aligned}
\Ex[
\RewardFunction{i}(\BigState{t},\BigAction{i}{t},\Action{-i}{t})
\mid
\ConditionalState{i}{t},\Policy{i}{t},\Action{-i}{t}]
&=
\sum_{\State{t},\History{i}{t}}
\ConditionalState{i}{t}(\State{t},\History{i}{t})
\sum_{\Action{i}{t}}
\Policy{i}{t}(\Action{i}{t}\mid\History{i}{t})
\RewardFunction{i}(\State{t},\Action{i}{t},\Action{-i}{t})
\\
&=
\RetCondRew{i}{t}(\ConditionalState{i}{t},\Policy{i}{t},\Action{-i}{t}).
\end{aligned}
\]
\end{proof}

\begin{lemma}
\label{lem:cond-kernel}
\(\TransitionCondKernel{i}{t}\colon \ConditionalSpace{i}{t} \times \PolicySpace{i}{t} \times \ActionSpace{-i}{}  \to \Simplex{\ConditionalSpace{i}{t+1}}\) is the pushforward of \(\RetCondObs{i}{t}\) by \(\RetCondTrans{i}{t}\).
\end{lemma}

\begin{proof}
Fix \((\ConditionalState{i}{t},\Policy{i}{t},\Action{-i}{t})\).
For every \(\ConditionalState{i}{t+1}\in\ConditionalSpace{i}{t+1}\),
\[
\begin{aligned}
&\Prob(
\BigConditionalState{i}{t+1}=\ConditionalState{i}{t+1}
\mid
\ConditionalState{i}{t},\Policy{i}{t},\Action{-i}{t})
\\
&=
\sum_{\PublicObservation{t+1},\Observation{-i}{t+1}}
\RetCondObs{i}{t}(\ConditionalState{i}{t},\Policy{i}{t},\Action{-i}{t},\PublicObservation{t+1},\Observation{-i}{t+1})
\cdot
\mathbf{1}\{
\ConditionalState{i}{t+1}
=
\RetCondTrans{i}{t}(\ConditionalState{i}{t},\Policy{i}{t},\Action{-i}{t},\PublicObservation{t+1},\Observation{-i}{t+1})
\}
\\
&=
\TransitionCondKernel{i}{t}(\ConditionalState{i}{t+1}\mid\ConditionalState{i}{t},\Policy{i}{t},\Action{-i}{t}).
\end{aligned}
\]
\end{proof}

For fixed \(\SuffixPolicy{i}{t}\), the recursion below defines \(\PrivValFunc{i}{k}{\SuffixPolicy{i}{t}}\colon \ConditionalSpace{i}{k}\to \mathbb{R}\) on \(\ConditionalSpace{i}{k}\) for every \(k\in\{t,\ldots,\Horizon\}\).
Define
\[
\textstyle \PrivValFunc{i}{k}{\SuffixPolicy{i}{t}}\colon \ConditionalState{i}{k} 
\mapsto \min_{\SuffixPolicy{-i}{t}} 
\Ex_{\ConditionalState{i}{k},\SuffixPolicy{i}{t}}[ \sum_{\kappa=k}^{\Horizon-1} 
\RewardFunction{i}(\BigState{\kappa},\BigAction{i}{\kappa},\BigAction{-i}{\kappa}) ].
\]


\begin{theorem}
\label{thm:conditional-dynamic-programme}
For fixed \(\SuffixPolicy{i}{t}\), \(\PrivValFunc{i}{k}{\SuffixPolicy{i}{t}}\) satisfies the conditional Bellman recursion for every \(k\ge t\).
\end{theorem}

\begin{proof}
At \(k=\Horizon\), set \(\PrivValFunc{i}{\Horizon}{\SuffixPolicy{i}{t}}\equiv0\).

Fix \(k<\Horizon\), \(\ConditionalState{i}{k}\in\ConditionalSpace{i}{k}\), and the rule \(\Policy{i}{k}\) induced by \(\SuffixPolicy{i}{t}\).

For every \(\Action{-i}{k}\in\ActionSpace{-i}{}\), Lemmas~\ref{lem:cond-reward} and~\ref{lem:cond-kernel} give
\[
\begin{aligned}
&\Ex[
\RewardFunction{i}(\BigState{k},\BigAction{i}{k},\Action{-i}{k})
+
\PrivValFunc{i}{k+1}{\SuffixPolicy{i}{t}}(\BigConditionalState{i}{k+1})
\mid
\ConditionalState{i}{k},\Policy{i}{k},\Action{-i}{k}]
\\
&=
\RetCondRew{i}{k}(\ConditionalState{i}{k},\Policy{i}{k},\Action{-i}{k})
+
\sum_{\ConditionalState{i}{k+1}\in\mathtt{supp}(\TransitionCondKernel{i}{k}
(\cdot\mid\ConditionalState{i}{k},\Policy{i}{k},\Action{-i}{k}))}
\TransitionCondKernel{i}{k}
(\ConditionalState{i}{k+1}\mid\ConditionalState{i}{k},\Policy{i}{k},\Action{-i}{k})
\PrivValFunc{i}{k+1}{\SuffixPolicy{i}{t}}(\ConditionalState{i}{k+1}).
\end{aligned}
\]
Taking \(\min_{\Action{-i}{k}\in\ActionSpace{-i}{}}\) gives the recursion.
\end{proof}

\subsection{Belief dynamics}
\label{app:belief-dynamics}

Define \(\RetBeliefRew{i}{t} \colon (\BeliefState{i}{t}, \JointPolicy{t}) \mapsto \Ex_{\BeliefState{i}{t}, \JointPolicy{t}}[ \RewardFunction{i}(\BigState{t}, \BigJointAction{t}) ]\). 
\begin{lemma}
\label{lem:belief-reward}
\(\RetBeliefRew{i}{t}\) is the expectation of \(\RetCondRew{i}{t}\) under \(b\).
\end{lemma}

\begin{proof}
Fix \(b\in\BeliefSpace{i}{t}\) and \(\JointPolicy{t}=(\Policy{i}{t},\Policy{-i}{t})\).

By definition,
\begin{align*}
\RetBeliefRew{i}{t}(b,\JointPolicy{t})
&\doteq 
\Ex_{\BeliefState{i}{t}, \JointPolicy{t}}[ \RewardFunction{i}(\BigState{t}, \BigJointAction{t}) ]
\\
&= 
\Ex_{\BeliefState{i}{t}, \JointPolicy{t}}
[ 
\Ex_{\BigConditionalState{i}{t}, \Policy{i}{t}, \Policy{-i}{t}(\BigConditionalState{i}{t})}[ \RewardFunction{i}(\BigState{t}, \BigJointAction{t}) ]
]
\\
&=
\sum_{\ConditionalState{i}{t}\in\ConditionalSpace{i}{t}}
b(\ConditionalState{i}{t})
\RetCondRew{i}{t}(
\ConditionalState{i}{t},
\Policy{i}{t},
\Policy{-i}{t}(\ConditionalState{i}{t})).
\end{align*}
\end{proof}

\begin{lemma}
\label{lem:belief-transition}
\(\RetBeliefTrans{i}{t}\) is the law of \(\BigConditionalState{i}{t+1}\) under \((b,\JointPolicy{t})\).
\end{lemma}

\begin{proof}
Fix \(b\in\BeliefSpace{i}{t}\) and \(\JointPolicy{t}=(\Policy{i}{t},\Policy{-i}{t})\).

For every \(\ConditionalState{i}{t+1}\in\ConditionalSpace{i}{t+1}\),
\[
\begin{aligned}
&\Prob_{b,\JointPolicy{t}}(\BigConditionalState{i}{t+1}=\ConditionalState{i}{t+1})
\\
&=
\sum_{\ConditionalState{i}{t}\in\ConditionalSpace{i}{t}}
b(\ConditionalState{i}{t})
\TransitionCondKernel{i}{t}
(
\ConditionalState{i}{t+1}
\mid
\ConditionalState{i}{t},
\Policy{i}{t},
\Policy{-i}{t}(\ConditionalState{i}{t}))
\\
&=
\RetBeliefTrans{i}{t}(b,\JointPolicy{t})(\ConditionalState{i}{t+1}).
\end{aligned}
\]
\end{proof}

\begin{lemma}
\label{lem:belief-min-decomposition}
For separable payoffs, the minimisation over \(\Policy{-i}{t}\) decomposes pointwise.
\end{lemma}

\begin{proof}
Let \(F\colon\ConditionalSpace{i}{t}\times\ActionSpace{-i}{}\to\RealSpace\).

For every \(\Policy{-i}{t}\),
\[
\sum_{\ConditionalState{i}{t}}b(\ConditionalState{i}{t})
F(\ConditionalState{i}{t},\Policy{-i}{t}(\ConditionalState{i}{t}))
\ge
\sum_{\ConditionalState{i}{t}}b(\ConditionalState{i}{t})
\min_{\Action{-i}{t}}F(\ConditionalState{i}{t},\Action{-i}{t}).
\]
Choose
\[
\Policy{-i}{t,*}(\ConditionalState{i}{t})
\in
\argmin_{\Action{-i}{t}\in\ActionSpace{-i}{}}
F(\ConditionalState{i}{t},\Action{-i}{t}).
\]
Then
\[
\sum_{\ConditionalState{i}{t}}b(\ConditionalState{i}{t})
F(\ConditionalState{i}{t},\Policy{-i}{t,*}(\ConditionalState{i}{t}))
=
\sum_{\ConditionalState{i}{t}}b(\ConditionalState{i}{t})
\min_{\Action{-i}{t}}F(\ConditionalState{i}{t},\Action{-i}{t}).
\]
\end{proof}



\subsection{Continuation values}
\label{app:continuation-values}

Define
\[
\ContVal{i}{t}(b)
\defeq
\max_{\SuffixPolicy{i}{t}}
\min_{\SuffixPolicy{-i}{t}}
\Ex_{b,\SuffixPolicy{i}{t},\SuffixPolicy{-i}{t}}
\left[
\sum_{k=t}^{\Horizon-1}
\RewardFunction{i}(\BigState{k},\BigJointAction{k})
\right].
\]

\begin{lemma}
\label{lem:contval-bellman}
\(\ContVal{i}{t}\) satisfies the focal belief Bellman recursion.
\end{lemma}

\begin{proof}
At \(t=\Horizon\), \(\ContVal{i}{\Horizon}=0\).

Fix \(t<\Horizon\).
Rearranging \(\max\) and \(\min\) operators require replacing opponent deterministic policy suffix \(\Policy{-i}{t}\) by a mixed one.
For any \((\Policy{i}{t}, \Policy{-i}{t})\), Lemmas~\ref{lem:belief-reward} and~\ref{lem:belief-transition} give
\[
\begin{aligned}
&\Ex_{b,\JointPolicy{t},\SuffixPolicy{i}{t+1},\SuffixPolicy{-i}{t+1}}
\left[\sum_{k=t}^{\Horizon-1}\RewardFunction{i}(\BigState{k},\BigJointAction{k})\right]
\\
&=
\RetBeliefRew{i}{t}(b,\JointPolicy{t})
+
\Ex_{\RetBeliefTrans{i}{t}(b,\JointPolicy{t}),\SuffixPolicy{i}{t+1},\SuffixPolicy{-i}{t+1}}
\left[\sum_{k=t+1}^{\Horizon-1}\RewardFunction{i}(\BigState{k},\BigJointAction{k})\right].
\end{aligned}
\]

Decomposing the policy space and optimising gives
\[
\ContVal{i}{t}(b)
=
\max_{\Policy{i}{t}}
\min_{\Policy{-i}{t}}
\left[
\RetBeliefRew{i}{t}(b,\JointPolicy{t})
+
\ContVal{i}{t+1}(\RetBeliefTrans{i}{t}(b,\JointPolicy{t}))
\right].
\]
where the interchange of $\max$ and $\min$ is established using arguments similar to those in \cite{DelBufDibSaf-DGAA-23} and \cite{Escudie2025}.
\end{proof}

\begin{theorem}
\label{thm:focal-belief-bellman}
The optimal focal belief value function \(\OptBeliefVal{i}{t}\) satisfies the focal belief Bellman equation and equals \(\ContVal{i}{t}\).
\end{theorem}

\begin{proof}
Both functions have terminal value zero.

Assume \(\OptBeliefVal{i}{t+1}=\ContVal{i}{t+1}\).
By Lemma~\ref{lem:contval-bellman},
\[
\ContVal{i}{t}(b)
=
\max_{\Policy{i}{t}}
\min_{\Policy{-i}{t}}
\left[
\RetBeliefRew{i}{t}(b,\JointPolicy{t})
+
\OptBeliefVal{i}{t+1}(\RetBeliefTrans{i}{t}(b,\JointPolicy{t}))
\right].
\]
The right-hand side is the Bellman definition of \(\OptBeliefVal{i}{t}(b)\).
Backward induction gives the claim.
\end{proof}

\subsection{Focal decomposition}
\label{app:focal-decomposition}


\begin{lemma}
\label{lem:prefix-belief-law}
A prefix pair \((\PrefixPolicy{i}{t},\PrefixPolicy{-i}{t})\) induces \(\BeliefState{i}{k+1}=\RetBeliefTrans{i}{k}(\BeliefState{i}{k},\JointPolicy{k})\).
\end{lemma}

\begin{proof}
At \(k=0\), \(\BeliefState{i}{0}\) is induced by \(\InitialBelief\).
If \(\BeliefState{i}{k}\) is the law of \(\BigConditionalState{i}{k}\), then Lemma~\ref{lem:belief-transition} gives
\[
\BeliefState{i}{k+1}
=
\RetBeliefTrans{i}{k}(\BeliefState{i}{k},\JointPolicy{k}).
\]
\end{proof}







\subsection{Proof of Theorem~\ref{thm:fixed_policy_linearity}}
\label{app:proof-value-structure}

For fixed \(\SuffixPolicy{i}{t}\), define
\[
\ValueFunctionErwan{}_{i,k}^{\SuffixPolicy{i}{t}}(b)
\defeq
\min_{\SuffixPolicy{-i}{k}}
\Ex_{b,\SuffixPolicy{i}{k},\SuffixPolicy{-i}{k}}
\left[
\sum_{\ell=k}^{\Horizon-1}
\RewardFunction{i}(\BigState{\ell},\BigJointAction{\ell})
\right],
\]
where \(\SuffixPolicy{i}{k}\) is the tail of \(\SuffixPolicy{i}{t}\).

\begin{lemma}
\label{lem:fixed-policy-linear}
For fixed \(\SuffixPolicy{i}{t}\) and \(k\ge t\), \(\ValueFunctionErwan{}_{i,k}^{\SuffixPolicy{i}{t}}\) is linear on \(\BeliefSpace{i}{k}\).
\end{lemma}

\begin{proof}
For \(c\in\ConditionalSpace{i}{k}\), set
\[
w_{i,k}^{\SuffixPolicy{i}{t}}(c)
\defeq
\PrivValFunc{i}{k}{\SuffixPolicy{i}{t}}(c).
\]

We prove by backward induction that
\[
\ValueFunctionErwan{}_{i,k}^{\SuffixPolicy{i}{t}}(b)
=
\sum_{c\in\ConditionalSpace{i}{k}}b(c)w_{i,k}^{\SuffixPolicy{i}{t}}(c).
\]
The claim is zero at \(k=\Horizon\).

Assume it holds at \(k+1\).
The stage-\(k\) focal rule \(\Policy{i}{k}\) is fixed.
By Lemmas~\ref{lem:belief-reward}, \ref{lem:belief-transition}, and the induction hypothesis,
\[
\begin{aligned}
\ValueFunctionErwan{}_{i,k}^{\SuffixPolicy{i}{t}}(b)
&=
\min_{\Policy{-i}{k}}
\sum_{c\in\ConditionalSpace{i}{k}}b(c)
\Biggl[
\RetCondRew{i}{k}(c,\Policy{i}{k},\Policy{-i}{k}(c))
\\
&\quad+
\sum_{c'\in\ConditionalSpace{i}{k+1}}
\TransitionCondKernel{i}{k}
(c'\mid c,\Policy{i}{k},\Policy{-i}{k}(c))
w_{i,k+1}^{\SuffixPolicy{i}{t}}(c')
\Biggr].
\end{aligned}
\]
By Lemma~\ref{lem:belief-min-decomposition},
\[
\begin{aligned}
\ValueFunctionErwan{}_{i,k}^{\SuffixPolicy{i}{t}}(b)
&=
\sum_{c\in\ConditionalSpace{i}{k}}b(c)
\min_{\Action{-i}{k}}
\Biggl[
\RetCondRew{i}{k}(c,\Policy{i}{k},\Action{-i}{k})
\\
&\quad+
\sum_{c'\in\ConditionalSpace{i}{k+1}}
\TransitionCondKernel{i}{k}
(c'\mid c,\Policy{i}{k},\Action{-i}{k})
w_{i,k+1}^{\SuffixPolicy{i}{t}}(c')
\Biggr].
\end{aligned}
\]
By Theorem~\ref{thm:conditional-dynamic-programme}, the inner minimum equals \(w_{i,k}^{\SuffixPolicy{i}{t}}(c)\).
Thus
\[
\ValueFunctionErwan{}_{i,k}^{\SuffixPolicy{i}{t}}(b)
=
\sum_{c\in\ConditionalSpace{i}{k}}b(c)w_{i,k}^{\SuffixPolicy{i}{t}}(c).
\]
Linearity follows.
\end{proof}

\begin{theorem}
\label{thm:convexity-focal-belief-value-proof}
\(\OptBeliefVal{i}{t}\) is convex, and \(\OptBeliefVal{i}{t}(b)=\max_{\ValueFunctionErwan{} \in \Gamma_{i,t}}\ValueFunctionErwan{}(b)\) for all \(b\in\BeliefSpace{i}{t}\).
\end{theorem}

\begin{proof}
Let
\[
\Gamma_{i,t}
\defeq
\{
\ValueFunctionErwan{}_{i,t}^{\SuffixPolicy{i}{t}}
:
\SuffixPolicy{i}{t}\in\SuffixPolicySpace{i}{t}
\}.
\]
By Lemma~\ref{lem:finite-compact-attainment}, \(\Gamma_{i,t}\) is compact in the finite-dimensional space of linear functionals.

By Theorem~\ref{thm:focal-belief-bellman},
\[
\OptBeliefVal{i}{t}(b)
=
\ContVal{i}{t}(b).
\]
By definition of \(\ContVal{i}{t}\),
\[
\ContVal{i}{t}(b)
=
\max_{\SuffixPolicy{i}{t}\in\SuffixPolicySpace{i}{t}}
\ValueFunctionErwan{}_{i,t}^{\SuffixPolicy{i}{t}}(b).
\]
Hence
\[
\OptBeliefVal{i}{t}(b)
=
\max_{\ValueFunctionErwan{}\in \Gamma_{i,t}}\ValueFunctionErwan{}(b).
\]
For \(b,b'\in\BeliefSpace{i}{t}\) and \(\lambda\in[0,1]\),
\[
\begin{aligned}
\OptBeliefVal{i}{t}(\lambda b+(1-\lambda)b')
&=
\max_{\ValueFunctionErwan{}\in \Gamma_{i,t}}
\ValueFunctionErwan{}(\lambda b+(1-\lambda)b')
\\
&=
\max_{\ValueFunctionErwan{}\in \Gamma_{i,t}}
\left[
\lambda\ValueFunctionErwan{}(b)+(1-\lambda)\ValueFunctionErwan{}(b')
\right]
\\
&\le
\lambda\OptBeliefVal{i}{t}(b)
+
(1-\lambda)\OptBeliefVal{i}{t}(b').
\end{aligned}
\]
Thus \(\OptBeliefVal{i}{t}\) is convex.
\end{proof}

\subsection{Local Bellman backups}
\label{app:proof-agpbpi-security}

The theorem is conditional on finite coverage, Bellman closure, globally valid inserted vectors, infinite exact updates, and covered-backup consistency.

\subsubsection{Local point-based backup}
\label{app:point-based-bellman-backups}

We introduce the following linear program to compute a greedy decision rule with respect to a fixed continuation policy $\SuffixPolicy{i}{t+1}$ and focal belief $\BeliefState{i}{t}$.

The program uses the following variables:
\begin{itemize}
    \item {\textcolor{black}{$\mathbf{v({\ConditionalState{i}{t})}}$}}, representing the value associated with conditional state $\ConditionalState{i}{t}$,
    \item \textcolor{black}{$\mathbf{\eta{(\History{-i}{t+1}, \History{0}{t+1})}}$}, representing auxiliary continuation values,
    \item \textcolor{black}{$\mathbf{\Policy{i}{t} \in \PolicySpace{i}{t}}$}, the focal decision rule to optimize.
\end{itemize}

\begin{theorem}
\label{thm:LP}
For any fixed suffix policy $\SuffixPolicy{i}{t+1}$ (associated with its value function $\Lambda_{i,t+1}$), and focal belief $\BeliefState{i}{t} \in \BeliefSpace{i}{t}$, solving the linear program
\begin{align*}
\textstyle
\text{Max} \hspace{0.5em}
& \textstyle \sum_{\ConditionalState{i}{t}}
\BeliefState{i}{t}(\ConditionalState{i}{t})\, \textcolor{black}{\mathbf{v{(\ConditionalState{i}{t})}}} \\
\text{Subject to} \hspace{0.5em}
& \textcolor{black}{\mathbf{v{(\ConditionalState{i}{t})}}}
\leq
\sum_{\Observation{-i}{t}, \Observation{0}{t}}
\textcolor{black}{\mathbf{\eta{(\History{-i}{t+1}, \History{0}{t+1})}}},
\quad
\forall \ConditionalState{i}{t}, \Action{-i}{t}
\\
& \textcolor{black}{\mathbf{\eta{(\History{-i}{t+1}, \History{0}{t+1})}}}
\leq
\sum_{\History{i}{t}, \Action{i}{t}}
\textcolor{black}{\mathbf{\Policy{i}{t}(\Action{i}{t} \mid \History{i}{t}, \History{0}{t})}}
\, Q_{(\History{-i}{t+1}, \History{i}{t+1}, \alpha)},
\\
& \hspace{16em}
\forall \History{-i}{t+1},
\History{0}{t+1},
\alpha \in \Lambda_{i,t+1},
\\
& \textcolor{black}{\mathbf{\Policy{i}{t}}} \in \PolicySpace{i}{t}
\end{align*}
where
\[
(\History{-i}{t+1}, \History{0}{t+1})
=
(\ConditionalState{i}{t},
\Action{-i}{t},
\Observation{-i}{t},
\Observation{0}{t}),
\]
yields a decision rule $\Policy{i}{t}$ that greedily maximizes
\[
\PessimisticQValue{\Policy{i}{t+1:\Horizon}}{t}
(\BeliefState{i}{t}, \Policy{i}{t})
=
\min_{\Policy{-i}{t}}
\RetBeliefRew{i}{t}
(\BeliefState{i}{t},
\Policy{i}{t},
\Policy{-i}{t})
+
\ValueFunctionErwan{}_{i,t+1}^{\SuffixPolicy{i}{t+1}}
\bigl(
\RetBeliefTrans{i}{t}
(
\BeliefState{i}{t},
\Policy{i}{t},
\Policy{-i}{t}
)
\bigr),
\]
corresponding to the Q-value induced by the fixed-policy continuation
$(\Policy{i}{t}, \SuffixPolicy{i}{t+1})$.
\end{theorem}

\begin{proof}
By construction, $\PessimisticQValue{\Policy{i}{t+1:\Horizon}}{t}$ is consistent with $\ValueFunctionErwan{}^{\SuffixPolicy{i}{t+1}}_{i,t}$, and an optimal $\Policy{-i}{t}$ can be taken deterministic (since the suffix policy $\SuffixPolicy{i}{t+1}$ is fixed).

A greedy decision rule maximizes $\PessimisticQValue{\Policy{i}{t+1:\Horizon}}{t}$ over reachable beliefs:
\begin{align*}
\max_{\Policy{i}{t}} \PessimisticQValue{\Policy{i}{t+1:\Horizon}}{t}(\BeliefState{i}{t}, \Policy{i}{t})
&= \max_{\Policy{i}{t}} \min_{\Policy{-i}{t}}
\Big[
\RetBeliefRew{i}{t}(\BeliefState{i}{t}, \Policy{i}{t}, \Policy{-i}{t})
+  \, \ValueFunctionErwan{}_{i,t}^{\SuffixPolicy{i}{t+1}}\!\left(\RetBeliefTrans{i}{t}(\BeliefState{i}{t}, \Policy{i}{t}, \Policy{-i}{t})\right)
\Big] \\
&= \max_{\Policy{i}{t}} \min_{\Policy{-i}{t}}
\Big[
\RetBeliefRew{i}{t}(\BeliefState{i}{t}, \Policy{i}{t}, \Policy{-i}{t})
+  \sum_{\ConditionalState{i}{t+1}} \BeliefState{i}{t+1}(\ConditionalState{i}{t+1})
\min_{\alpha^{\SuffixPolicy{i}{t+1}}}
\alpha^{\SuffixPolicy{i}{t+1}} \cdot \ConditionalState{i}{t+1}
\Big],
\end{align*}
where $\BeliefState{i}{t+1} = \RetBeliefTrans{i}{t}(\BeliefState{i}{t}, \Policy{i}{t}, \Policy{-i}{t})$.

Expanding the belief update and rearranging terms, we obtain:
\begin{align*}
\max_{\Policy{i}{t}} \PessimisticQValue{\Policy{i}{t+1:\Horizon}}{t}(\BeliefState{i}{t}, \Policy{i}{t}) 
&= \max_{\Policy{i}{t}} \min_{\Policy{-i}{t}}
\sum_{\History{-i}{t}, \Action{-i}{t}, \Observation{-i}{t}, \Observation{0}{t}}
\Policy{-i}{t}(\Action{-i}{t} \mid \History{-i}{t}, \History{0}{t}) \\
&\quad \cdot \min_{\alpha^{\SuffixPolicy{i}{t+1}}}
\sum_{\History{i}{t}, \Action{i}{t}}
\Policy{i}{t}(\Action{i}{t} \mid \History{i}{t}, \History{0}{t})
\sum_{\State{t}} \BeliefState{i}{t}(\State{t}, h_t) \\
&\quad \cdot \Big[
\RetBeliefRew{i}{t}(\State{t}, \Action{i}{t}, \Action{-i}{t})
+  \sum_{\State{t+1}, z_t}
p(\State{t+1}, z_t, \Observation{0}{t} \mid \State{t}, a_t)
\alpha^{\SuffixPolicy{i}{t+1}}(\ConditionalState{i}{t+1})
\Big].
\end{align*}

This expression makes explicit the max–min structure together with the inner minimization over $\alpha$, where $\alpha$ is taken after the expectation over next-stage variables. This placement is crucial: it preserves the piecewise-linear concave structure of the value function and yields directly the desired linear programming formulation. 

If we denote
\(
(\History{-i}{t+1}, \History{0}{t+1}) = (\ConditionalState{i}{t}, \Action{-i}{t}, \Observation{-i}{t}, \Observation{0}{t})
\), using the bijection between $\ConditionalState{i}{t}$ and the pair of histories $(\History{-i}{t}, \History{0}{t})$ 
, we can define
\begin{align*}
&Q_{(\History{-i}{t+1}, \History{i}{t+1}, \alpha)}\\
&= \sum_{\State{t}} \BeliefState{i}{t}(\State{t}, h_t)
\Bigg[
\RetBeliefRew{i}{t}(\State{t}, \Action{i}{t}, \Action{-i}{t})
+  \sum_{\State{t+1}, z_t}
p(\State{t+1}, z_t, \Observation{0}{t} \mid \State{t}, a_t)\,
\alpha^{\SuffixPolicy{i}{t+1}}(\ConditionalState{i}{t+1})
\Bigg],    
\end{align*}
then the above quantity corresponds to the $q$-value of the best response where player $-i$ plays $\Action{-i}{t}$ in $\ConditionalState{i}{t}$ and follows $\alpha^{\SuffixPolicy{i}{t+1}}$ at $(\History{-i}{t+1}, \History{0}{t})$. This yields the coefficients of the linear program described in the theorem.

We also impose an additional constraint on $\Policy{i}{t}$ to ensure that it is a valid decision rule, i.e., that each local strategy belongs to the simplex:
\[
\Policy{i}{t}(\cdot \mid \History{i}{t}, \History{0}{t}) \in \Simplex{\ActionSpace{i}{}}, \quad \forall \History{i}{t}, \History{0}{t}
\]
ensuring that $\Policy{i}{t}$ defines a proper probability distribution over actions at every local history.
\end{proof}


\label{appendix:LP:complexity}

\begin{table}[t]
\centering
\small
\setlength{\tabcolsep}{10pt}
\renewcommand{\arraystretch}{1.}
\begin{tabular}{p{2.cm} c c}
\toprule
\textbf{Update step} & \textbf{\# variables} & \textbf{\# constraints} \\
\midrule

\textbf{Ours: LP} &
$|\BigHistory{1}{t}||\BigAction{1}{t}|+|\BigHistory{2}{t}||\BigAction{2}{t}||\BigObservation{2}{t}| + |\BigHistory{2}{t}|$ &
$O(|\BigHistory{2}{t}||\BigAction{2}{t}||\BigObservation{2}{t}||\Lambda|)$ \\

\midrule

\citep{DelBufDibSaf-DGAA-23} &
(rule variables $+$ epigraph) &
\textbf{Exponential:}\;\; $O\!\bigl(|\BigAction{2}{t}|^{|\BigHistory{2}{t}|}\bigr)$ \\

\citep{Escudie2025} &
$|\BigHistory{1}{t}||\BigAction{1}{t}||\Gamma_{1}|+|\BigHistory{2}{t}||\BigAction{2}{t}||\BigObservation{2}{t}| + |\BigHistory{2}{t}|$ &
$O\!\bigl(|\Gamma_{1}|\cdot|\Gamma^{*}_{2}|\cdot|\BigHistory{2}{t}|\cdot|\BigAction{2}{t}|\cdot|\BigObservation{2}{t}|\bigr)$ \\
\bottomrule
\end{tabular}
\caption{LP complexity comparison (input-size view).
The OMG greedy-update LP has constraints indexed by opponent decision rules, yielding an exponential dependence on $|\BigHistory{2}{t}|$ \citep{DelBufDibSaf-DGAA-23}.
Both the cached-envelope greedy LP \citep{Escudie2025} and our backup scale polynomially in the same explicit representation objects, but expose different optimization structures. In our formulation, both the number of variables and constraints are reduced by a factor of $|\Gamma_1|$. Here, $|\Gamma_1|$ denotes the number of piecewise-linear concave functions maintained, while $|\Lambda|$ is the number of piecewise-linear vectors associated with each element of $\Gamma_1$.}
\label{tab:lp-complexity-comparison}
\end{table}

\subsection{Comparative Analysis of Backup Complexity}
\label{app:LP:PBVIvsAPI}
\paragraph{Comparative analysis: \citet{DelBufDibSaf-DGAA-23}}
The OMG greedy-update formulation \citet{DelBufDibSaf-DGAA-23} enforces pessimism against \emph{all} deterministic opponent decision rules
$\Policy{2}{t}:\BigHistory{2}{t}\to \BigAction{2}{t}$.
Because there are $|\BigAction{2}{t}|^{|\BigHistory{2}{t}|}$ such mappings, the LP description contains
$O(|\BigAction{2}{t}|^{|\BigHistory{2}{t}|})$ constraints (Table~\ref{tab:lp-complexity-comparison}), which is exponential in the size of the opponent’s private-history space.
Our backup removes this bottleneck by never indexing constraints by opponent decision rules:
our LP constrains only against finitely many vectors in a cached envelope $\Lambda$. Hence our LP admits polynomial-size descriptions in the explicit parameters
$|\BigHistory{1}{t}|,|\BigHistory{2}{t}|,|\BigAction{1}{t}|,|\BigAction{2}{t}|,|\BigObservation{2}{t}|,|\Lambda|$.

\paragraph{Comparative analysis: \citet{Escudie2025}}
In \citep{Escudie2025}, the LP explicitly incorporates all candidate next policies of $i$ through the sets $\Gamma_1$ of piecewise linear concave function $\Lambda$, which grow over iterations. Specifically, a single monolithic greedy LP jointly couples the outer cache index $\Gamma_1$, the inner value index $\alpha \in \Lambda$, and the opponent action-observation indices $(\Action{2}{t}, \Observation{2}{t})$. 
This full coupling within a single program leads to a constraint count of 
\(
O\big(|\Gamma_1|\cdot|\Lambda|\cdot|\BigHistory{2}{t}|\cdot|\BigAction{2}{t}|\cdot|\BigObservation{2}{t}|\big)
\),
as reported in Table~\ref{tab:lp-complexity-comparison}. 

Although the reduction in $\mid \Gamma_1 \mid$ has little impact for short horizons, it becomes critical for longer horizons. Indeed, in PBVI \citep{Escudie2025}, the size of $\Gamma_1$ grows with the number of sampled points, which itself increases with the number of stages required to cover an ever-expanding continuous state space.

\section{Algorithm}

Value iteration in \citep{Escudie2025} amounts to solving a focal belief game using sampled opponent behaviors. However, this unguided sampling process induces an exponential growth in the number of sampled points, resulting in a rapidly increasing computational complexity. To address this limitation, we introduce a policy iteration approach that exploits the structural properties of the game together with the reduced formulations defined previously. Leveraging this structure yields a more efficient iterative procedure while preserving optimality guarantees.

We prove that the dynamics induced by the game can be viewed as a discretization of a continuous-time dynamical system. Along the resulting continuous trajectory, the dual gap is non-increasing and converges to zero. The proof relies on three key properties: the convexity of the dual gap in the policy space, the differentiability of the value function almost everywhere, and the uniform continuity of the value function. Together, these properties allow us to characterize the evolution of the dual gap along the continuous dynamics and establish its convergence.

This section introduces and formally defines the main objects used throughout the analysis, including the dual gap and the dynamics guiding the update $\Dynamics{i}{t}$, and related quantities.

The continuous dynamics describe the infinitesimal policy updates underlying the algorithm. At each stage $t$, the update acts on the policy $\Policy{i}{t}$ while the policies at other stages are kept fixed. The resulting direction therefore belongs to the local policy space $\PolicySpace{i}{t}$, while depending on the complete policy of player $i$ and on the opponent's policies up to stage $t-1$ through the induced belief state. This defines a vector field over the policy space, whose integral curves describe the continuous version of the algorithm. 

\begin{definition}[Continuous Dynamics]
\label{def:continuous_dynamics}
For player $i$ and stage $t$, the continuous dynamics are defined as the
vector field
\[
\Dynamics{i}{t} :
\PolicySpace{i}{0:\Horizon-1}
\times
\PolicySpace{-i}{0:t-1}
\rightarrow
\PolicySpace{i}{t},
\]
given by
\[
\Dynamics{i}{t}
\left(
\Policy{i}{0:\Horizon-1},
\Policy{-i}{0:t-1}
\right)
=
\nabla_{\Policy{i}{t}}
\PessimisticQValue{\Policy{i}{t+1:\Horizon}}{}
\left(
\BeliefState{i}{t},\Policy{i}{t}
\right),
\]

where $\BeliefState{i}{t}$ is the focal belief induced by $\Policy{i}{0:t-1}$ and $\Policy{-i}{0:t-1}$ and $\PessimisticQValue{\Policy{i}{t+1:\Horizon-1}}{}$ is defined in Theorem~\ref{thm:LP}. Thus, $\Dynamics{i}{t}$ gives the safest most improving partial direction of the pessimistic $Q$-value with respect to $\Policy{i}{t}$, while keeping the belief fixed.
\end{definition}

We may ask whether having the agents follow their respective induced flows improves their value $\ValueFunctionErwan{}^i_t$. The direction induced by $\Dynamics{i}{t}$ is safely improving for timestep $t$ to $\Horizon$, as all future best-responses are considered in $\PessimisticQValue{\Policy{i}{t+1:\Horizon-1}}{}$. However, it remains unclear whether this improvement is preserved globally across all stages. Indeed, $\Dynamics{i}{t}$ is defined under the assumption that the opponent's past policy remains fixed, as reflected by its dependence on $\Policy{-i}{0:t-1}$. When the policies are updated simultaneously across iterations, this assumption may no longer hold: changes to the opponent's policy at earlier stages can modify the induced belief $\BeliefState{i}{t}$ and, consequently, the continuation value at later stages. This creates a situation similar to fictitious play, where both players update their policies simultaneously and each player's update can therefore affect the optimization problem faced by the other. 

\begin{definition}[Dual gap]
Given a joint policy $\Policy{}{t:\Horizon-1}$ and the belief state $\BeliefState{i}{t}$ induced by the joint policy up to time $t-1$, the dual gap at time $t$ is defined as
\[
    \DualGap{t}^{\Policy{}{t:\Horizon-1}}
    \left(
        \BeliefState{i}{t}
    \right)
    =
    \PessimisticQValue{\Policy{-i}{t+1:\Horizon-1}}{t}
    \left(
        \BeliefState{i}{t},
        \Policy{-i}{t}
    \right)
    -
    \PessimisticQValue{\Policy{i}{t+1:\Horizon-1}}{t}
    \left(
        \BeliefState{i}{t},
        \Policy{i}{t}
    \right).
\]
\end{definition}

\begin{lemma}[Local stability of the best response]
\label{lem:local_br}
At any regular policy profile, the best response of the opponent in the constrained subgame rooted at the current policy is locally constant. Consequently, the pessimistic $Q$-value
\[
    {\PessimisticQValue{i}{t}}
    \left(
        \BeliefState{i}{t},\Policy{i}{t}
    \right) \dot{=} \PessimisticQValue{i,\Policy{i}{t+1:\Horizon-1}}{t}
    \left(
        \BeliefState{i}{t},\Policy{i}{t}
    \right)
\]
defined in Theorem~\ref{thm:LP} is differentiable with respect to $\Policy{i}{t}$ in a sufficiently small neighborhood of the current policy.
\end{lemma}

\begin{proof}
At a regular policy profile, the opponent's best response is unique, since the game is concave-convex and admits a finite set of deterministic policies. Consequently, there exists a neighborhood of the current policy in which this best response remains unchanged. Within this neighborhood, the pessimistic continuation value is therefore evaluated against a fixed best response. The remaining terms in the Bellman recursion are differentiable with respect to $\Policy{i}{t}$. Hence,

$$
    \PessimisticQValue{i}{t}
    \left(
        \BeliefState{i}{t},\Policy{i}{t}
    \right)
$$

is differentiable with respect to $\Policy{i}{t}$.
\end{proof}

The Q-value function is not differentiable everywhere, but this issue is not specific to our setting: the same non-differentiability arises for the dual gap in normal-form games. Since our method uses the same type of local improvement mechanism in policy space, it can handle these points in the same way. Moreover, our discretization does not encounter pathological cases in practice, and the value function explicitly accounts for possible deviations from the current policy. This provides additional protection against numerical instabilities caused by nondifferentiable points and ensures that the improvement step remains robust to such cases.

\begin{lemma}[Local improving direction]
\label{lem:local_improvement}
At a regular policy profile, the direction $\Dynamics{i}{t}$ is an improving direction for player $i$'s pessimistic value at stage $t$. In particular,

$$
    \left.
    \frac{d}{d\alpha}
    \PessimisticQValue{i}{t}
    \left(
        \BeliefState{i}{t},
        \Policy{i}{t}
        +
        \alpha\Dynamics{i}{t}
    \right)
    \right|_{\alpha=0}
    >0
$$

whenever $\Policy{i}{t}$ is not locally optimal.
\end{lemma}

\begin{proof}
By Lemma~\ref{lem:local_br}, the local pessimistic $Q$-value is differentiable around the current policy. By construction of $\Dynamics{i}{t}$,

$$
    \Dynamics{i}{t}
    =
    \nabla_{\Policy{i}{t}}
    \PessimisticQValue{i}{t}
    \left(
        \BeliefState{i}{t},\Policy{i}{t}
    \right).
$$

Therefore,

$$
    \left.
    \frac{d}{d\alpha}
    \PessimisticQValue{i}{t}
    \left(
        \BeliefState{i}{t},
        \Policy{i}{t}
        +
        \alpha\Dynamics{i}{t}
    \right)
    \right|_{\alpha=0}
    =
    \left\|
        \Dynamics{i}{t}
    \right\|^2
    >0
$$

whenever $\Dynamics{i}{t}\neq 0$.
\end{proof}

\begin{lemma}[Safe improvement for $\Horizon=2$]
\label{lem:two_stage_safe_improvement}
Suppose that $\Horizon=2$. For sufficiently small step sizes, the update of player $i$ following the dynamics $\Dynamics{i}{t}$ strictly improves her pessimistic value whenever the current policy is not locally optimal.
\end{lemma}

\begin{proof}
Consider the joint policy $(\Policy{1,k}{0:1},\Policy{2,k}{0:1})$ at iteration $k$. The decision rule $\Policy{i,k}{0}$ is optimal for the pessimistic continuation problem induced by $\Policy{i,k}{1}$.

We first update the last-stage decision rule according to the above dynamics. Since there is no subsequent decision rule, the associated one-stage pessimistic problem is solved directly. Thus, considering any step size $\alpha_1^k>0$, the update

\[
    \Policy{i,k+1}{1}
    =
    \Policy{i,k}{1}
    +
    \alpha_1^k
    \Dynamics{i}{1}
\]

improves the corresponding local pessimistic value, and hence decreases the corresponding local dual gap.

We then update the first-stage decision rule while keeping the newly
obtained $\Policy{i,k+1}{1}$ fixed. Define

$$
    v^k
    =
    \PessimisticValue{i, [\Policy{i,k}{0},
        \Policy{i,k}{1}]}{0},
$$

$$
    \widetilde v^k
    =
    \PessimisticValue{i, [\Policy{i,k}{0},
        \Policy{i,k+1}{1}]}{0},
$$

and

\[
    v^{k+1}
    =
    \PessimisticValue{i, [\Policy{i,k+1}{0},
        \Policy{i,k+1}{1}]}{0}.
\]

By construction of the first-stage update, $\Dynamics{i}{0}$ is an improving direction for the local function. Therefore, the newly obtained policy $\Policy{i,k+1}{0}$ strictly improve the overall pessimistic value

\[
    \Policy{i}{0}
    \mapsto
    \PessimisticValue{i, [\Policy{i}{0},
        \Policy{i,k+1}{1}]}{0},
\]

and thus,

\[
    v^{k+1}>\widetilde v^k.
\]

It remains to ensure that the improvement at stage $t=0$ is not cancelled by the preceding update at stage $t=1$. To this end, we regard the stage $t=1$ update as a homotopy

\[
    \Policy{i,k+1}{1}(\alpha)
    =
    \Policy{i,k}{1}
    +
    \alpha
    \Dynamics{i}{1},
    \qquad
    \alpha\in[0,\bar\alpha].
\]

Let

\[
    \Delta(\alpha)
    =
    \PessimisticValue{i, [\Policy{i,k+1}{0}(\alpha),
        \Policy{i,k+1}{1}(\alpha)]}{0}
    -
    \PessimisticValue{i, [\Policy{i,k}{0},
        \Policy{i,k}{1}]}{0},
\]

where $\Policy{i,k+1}{0}(\alpha)$ denotes the stage $t=0$ update computed from the continuation policy $\Policy{i,k+1}{1}(\alpha)$.

At $\alpha=0$, the continuation policy coincides with the policy at iteration $k$. Since $\Policy{i,k+1}{0}$ is an improving direction for the corresponding local problem, we have $\Delta(0)>0$ at every non-degenerate point where the stage $0$ direction is strictly improving. By continuity of the pessimistic value in the policy and, locally, of the stage $0$ update, there exists $\bar\alpha>0$ such that

\[
    \Delta(\alpha)>0
    \qquad
    \text{for all }0<\alpha<\bar\alpha.
\]

Hence, the stage $1$ step can be chosen sufficiently small so that the
improvement obtained at stage $0$ is not cancelled by the preceding
continuation update. Consequently,

$$
    v^{k+1}>v^k.
$$

\end{proof}

\begin{lemma}
\label{lem:safe_improvement}
For a horizon $\Horizon$, if the stage wise updates from stage $1$ to stage $\Horizon-1$ can be chosen to strictly improve the pessimistic value, then the update at stage $t=0$ can also be chosen so that the complete iteration strictly improves the pessimistic value.
\end{lemma}

\begin{proof}
Assume that the result holds for a horizon $\Horizon$. Consider a horizon $\Horizon+1$. Starting from the last stage, we apply the construction from Lemma~\ref{lem:two_stage_safe_improvement} to stages

\[
    1,\ldots,\Horizon-1.
\]

By the induction hypothesis, the continuation problem from stage $1$ onwards can be updated with sufficiently small step sizes so that its pessimistic value is improved. 

We then consider stage $t=0$ while keeping the resulting continuation policies fixed. The same local argument applies: the continuation pessimistic value induces a differentiable local $Q$-value around the current stage $t=0$ policy, and the update follows an improving direction of this function.

Finally, the step sizes at stages $1,\ldots,\Horizon-1$ can be chosen sufficiently small so that the improvement produced at stage $0$ dominates the change induced by the continuation updates. By continuity, there therefore exist positive step sizes

\[
    \alpha_0^k,\ldots,\alpha_{\Horizon}^k
\]

for which the complete iteration yields a strict improvement of the
pessimistic value.
\end{proof}


\begin{lemma}
\label{lem:dual_gap_decreasing}
For any starting joint policy $\Policy{}{}$, there exists a discretized trajectory

\[
\phi = (\Policy{1,k}{}, \Policy{2,k}{})_{k\geq 0}
\]

starting from any $(\Policy{1,0}{}, \Policy{2,0}{})$ and following the flow induced by $(\Dynamics{i}{0:\Horizon-1},\Dynamics{-i}{0:\Horizon-1})$ such that the dual gap strictly decreases at every non-degenerate iteration.
\end{lemma}

\begin{proof}
We construct the trajectory iteratively. At iteration $k$, the policy at each stage is updated according to
\[
\Policy{i,k+1}{t}
=
\Policy{i,k}{t}
+
\alpha_t^k
\Dynamics{i}{t}
\left(
\Policy{i,k}{0:\Horizon-1},
\Policy{-i,k}{0:\Horizon-1}
\right),
\]
with $\alpha_t^k>0$.

By Lemma~\ref{lem:safe_improvement}, whenever an improvement at an earlier stage could be affected by the updates of the continuation stages, the corresponding continuation step sizes can be chosen sufficiently small so that this improvement is preserved.

Consequently, at every non-degenerate iteration, the step sizes
\[
\alpha_0^k,\ldots,\alpha_{\Horizon-1}^k
\]
can be reduced arbitrarily while remaining strictly positive and can be chosen so that the complete iteration strictly improves the pessimistic value. Equivalently, the associated dual gap satisfies
\[
\DualGap{0}(\BeliefState{}{0}, (\Policy{1,k+1}{}, \Policy{2,k+1}{}))
<
\DualGap{0}(\BeliefState{}{0}, (\Policy{1,k}{}, \Policy{2,k}{})).
\]

Starting from any $\Policy{}{}$, choosing such step sizes at each iteration defines a discretized trajectory $\phi$ following the induced flow. Along this trajectory, the dual gap therefore strictly decreases at every non-degenerate iteration.
\end{proof}

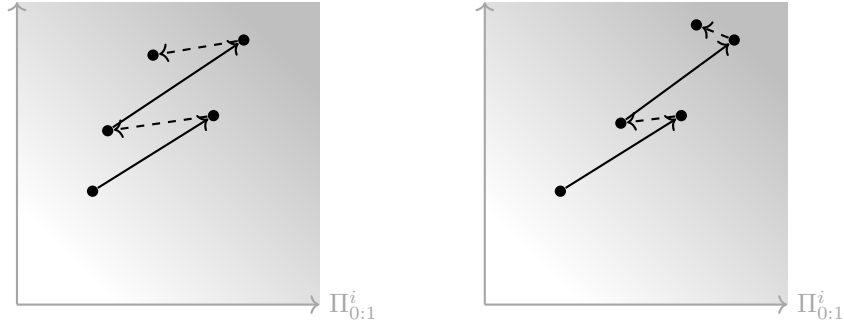
\begin{figure}[h]
\centering
\begin{tikzpicture}[
    point/.style={circle, fill=black, inner sep=1.5pt},
    fleche_noire/.style={->, thick, color=black},
    fleche_pointillee/.style={->, thick, color=black, dashed},
    axe/.style={->, thick, color=gray!70}
]

\begin{scope}[on background layer]
\shade[
        left color=gray!50,
        right color=gray!0,
        shading angle=-45
    ] (0, 0) rectangle (4., 4.);
\end{scope}

\draw[axe] (0, 0) -- (4., 0) node[right] {$\PolicySpace{i}{0:1}$};
\draw[axe] (0, 0) -- (0, 4.) node[above] {};

\node[point, label=below:] (p1) at (2*1-1, 2*1.5-1.5) {};
\node[point, label=below:] (p2) at (2*1.8-1, 2*2.-1.5) {};
\node[point, label=below:] (p3) at (2*1.1-1, 2*1.9-1.5) {};
\node[point, label=below:] (p4) at (2*2.-1, 2*2.5-1.5) {};
\node[point, label=below:] (p5) at (2*1.4-1, 2*2.4-1.5) {};

\draw[fleche_noire] (p1) -- (p2); 
\draw[fleche_pointillee] (p2) -- (p3); 
\draw[fleche_noire] (p3) -- (p4); 
\draw[fleche_pointillee] (p4) -- (p5); 

\end{tikzpicture}
\hspace{1cm}
\begin{tikzpicture}[
    point/.style={circle, fill=black, inner sep=1.5pt},
    fleche_noire/.style={->, thick, color=black},
    fleche_pointillee/.style={->, thick, color=black, dashed},
    axe/.style={->, thick, color=gray!70}
]
\begin{scope}[on background layer]
\shade[
        left color=gray!50,
        right color=gray!0,
        shading angle=-45
    ] (0, 0) rectangle (4., 4.);
\end{scope}

\draw[axe] (0, 0) -- (4., 0) node[right] {$\PolicySpace{i}{0:1}$};
\draw[axe] (0, 0) -- (0, 4.) node[above] {};

\node[point, label=below:] (p1) at (2*1-1, 2*1.5-1.5) {};
\node[point, label=below:] (p2) at (2*1.8-1, 2*2.-1.5) {};
\node[point, label=below:] (p3) at (2*1.1-0.4, 2*1.9-1.4) {};
\node[point, label=below:] (p4) at (2*2.-0.7, 2*2.5-1.5) {};
\node[point, label=below:] (p5) at (2*1.4, 2*2.4-1.1) {};

\draw[fleche_noire] (p1) -- (p2); 
\draw[fleche_pointillee] (p2) -- (p3); 
\draw[fleche_noire] (p3) -- (p4); 
\draw[fleche_pointillee] (p4) -- (p5); 

\end{tikzpicture}

\caption{\textbf{Example of a situation at $\Horizon = 2$}. The background represents the gradient of the dual gap at $t=0$. \textit{The black arrows} represent updates of the policy where only $\Policy{i}{0}$ is changing. At $t=0$, since $\Player{i}$ has an optimal view of the pessimistic value function, the move strictly improves the global dual gap. \textit{The dashed arrows} represent updates at $t=1$; since the update is also influenced by $\Player{-i}$ past, nothing guarantees that the update will follow the global dual gap $\DualGap{0}$ gradient direction. Here, we present an extreme case where the update at $t=1$ almost cancels the one at $t=0$. \textbf{On the left}, we do not adapt the step sizes, \textbf{on the right}, we apply the proof such that destructive updates are avoided and we are sure the updates go to the right direction.}
\end{figure}

We have established the existence of a trajectory following the flow induced by $\Dynamics{i}{t}$. Since the step size can be chosen arbitrarily small, we can construct a sequence of discrete trajectories with vanishing step sizes that increasingly closely approximates the continuous flow. Along this trajectory, the dual gap $\DualGap{0}$ is non-increasing. It remains to show that the trajectory converges to the set of Nash equilibria.

\begin{lemma}
\label{lem:inductive_improvement}
Suppose that, at a joint policy $\Policy{}{}$, a safe local improving direction exists for the overall objective. Then there exists a player $i$ and a stage $t$ such that the dynamics $\Dynamics{i}{t}$ identify a strictly improving direction at the corresponding induced belief state $\BeliefState{i}{t}$.
\end{lemma}

\begin{proof}
We prove the result by contradiction. Consider first the case $\Horizon=2$. Suppose that a safe improving direction exists (considering all stages at once), but that no improving direction can be identified inductively at either stage.

In particular, suppose that no improving direction can be identified at $t=0$. Then the improvement cannot be obtained from the current decision alone and must arise from the continuation decision at $t=1$. Hence, there exists an induced belief state $\BeliefState{i}{1}$, arbitrarily close to the current one, at which a local change of the policy at $t=1$ strictly improves the continuation objective.

By the local improvement property at $t=1$, this strict improvement implies the existence of an improving direction at that belief state. Since the continuation value is continuous with respect to the induced belief state, the same improving direction remains improving in a neighborhood of the current belief state. It can therefore be detected at the current belief state, contradicting the assumption that no improving direction can be identified inductively.

Thus, for $\Horizon=2$, every safe local improvement is detected at one of the two stages.

The argument extends inductively to an arbitrary finite horizon. Suppose that no improving direction can be identified at any stage. Then, after ruling out an improvement at the current stage, any remaining improvement must arise from the continuation problem. Applying the same argument to this continuation problem yields a later stage at which a strict local improvement exists. By continuity, this improvement must already be detectable at the corresponding current induced belief state, contradicting the assumption.

Therefore, whenever a safe global improving direction exists, the dynamics identify a strictly improving direction at some stage $t$.
\end{proof}

The next theorem characterizes the points toward which the continuous dynamics converges. 

\begin{theorem}
\label{thm:dual_gap_convergence}
The continuous dynamics $\phi$ introduced in Lemma~\ref{lem:dual_gap_decreasing} converges to the set of Nash equilibria of the game.
\end{theorem}

\begin{proof}
Consider a joint policy

\[
    \Policy{}{}\notin\mathcal{N},
\]

outside of the set of the Nash Equilibria. Since the dual gap vanishes if and only if the joint policy is a Nash equilibrium,

\[
    \DualGap{0}(\Policy{}{})>0.
\]

\[
    D_{\Dynamics{i}{t}}
    \DualGap{0}(\Policy{}{})
    <0.
\]

Therefore, at every joint policy outside $\mathcal{N}$, the dynamics admit a direction that strictly decreases the dual gap. By Lemma~\ref{lem:dual_gap_decreasing}, the updates can be chosen sufficiently small so that the simultaneous evolution of the two players does not cancel this strict improvement. Consequently, along the continuous dynamics,

$$
    \frac{\mathrm{d}}{\mathrm{d}\tau}
    \DualGap{0}(\phi(\tau))
    <0
    \qquad
    \text{whenever }
    \phi(\tau)\notin\mathcal{N}.
$$

Thus, the dual gap is a strict Lyapunov function outside the set $\mathcal{N}$.

Because the joint policy space is compact and the dual gap is non-negative, the function

$$
    g(\tau)
    =
    \DualGap{0}(\phi(\tau))
$$

is monotonically decreasing and bounded from below. Hence, it converges to some $g_\infty\geq0$.

Suppose, for contradiction, that

$$
    g_\infty>0.
$$

Then $\phi(\tau)$ remains outside $\mathcal{N}$ for all sufficiently large $\tau$. At every such point, the argument above provides a strictly improving direction for the dynamics, and therefore the dual gap must continue to decrease. This contradicts the convergence of $g(\tau)$ to the strictly positive limit $g_\infty$.

Hence,

$$
    g_\infty=0,
$$

and therefore

$$
    \lim_{\tau\rightarrow\infty}
    \DualGap{0}(\phi(\tau))
    =0.
$$

Since the dual gap is zero if and only if the joint policy belongs to $\mathcal{N}$, the trajectory $\phi(\tau)$ converges to the set of Nash equilibria.
\end{proof}

The previous result establishes that the continuous dynamics converge to the set of Nash equilibria. We now relate these continuous trajectories to the discrete-time policy updates considered in practice. In particular, we show that, under the appropriate conditions, the discrete dynamics track the continuous dynamics sufficiently closely to inherit the same convergence property. This connection is formalized in the following theorem.

\begin{proof}[Proof of Theorem~\ref{thm:algorithm_convergence}]
The discrete algorithm can be written as a stochastic approximation of the continuous dynamics introduced in Definition~\ref{def:continuous_dynamics}. In particular, for each player $i$,
\[
    \Policy{i,k+1}{}
    =
    (1-\alpha_k) \Policy{i,k}{}
    +
    \alpha_k
    \tilde{\Dynamics{i}{t}}
    \left(
        \Policy{i,k}{},
        \Policy{-i,k}{}
    \right)
    +
    U_{i,k+1}.
\]
The operator $\tilde{\Dynamics{i}{t}}$ is the discrete counterpart of the continuous improving direction $\Dynamics{i}{t}$. Both operators perform the same local pessimistic optimization, but they differ in how this optimization is carried out. While $\Dynamics{i}{t}$ describes the continuous direction of improvement, $\tilde{\Dynamics{i}{t}}$ is obtained by solving the stage-wise LPs used by the algorithm. These LPs exploit the uniform continuity of the value function to construct a local approximation of the pessimistic $Q$-value. In particular, $\tilde{\Dynamics{i}{t}}$ maximizes this local approximation while matching the exact pessimistic value at the current point (see Theorem~\ref{thm:LP}). Thus, the LP provides a discrete approximation of the same local optimization characterized by $\Dynamics{i}{t}$.

Consequently, the interpolated sequence of policies can be viewed as a stochastic approximation of the continuous system
\[
    \dot{\Policy{i}{}}
    =
    \Dynamics{i}{}
    \left(
        \Policy{i}{},
        \Policy{-i}{}
    \right).
\]
Under the Robbins-Monro conditions and the standard assumptions on the noise, the cumulative effect of the stochastic perturbations remains controlled, while the updates continue to explore the policy space. Together with the assumed consistency of the LP operator with the local continuous dynamics, this ensures that the interpolated trajectory is an asymptotic pseudo-trajectory of the flow induced by $\left(\Dynamics{1}{},\Dynamics{2}{}\right)$.

Proposition 1.3 of \cite{benaim2005} provides precisely the required connection. It states, under the corresponding stochastic approximation conditions, that the asymptotic behavior of the algorithm is preserved when its iterates are viewed as perturbed solutions of the limiting differential inclusion. Since we have established that the trajectories of this continuous dynamics converge to a Nash equilibrium, the proposition allows us to transfer this convergence result to the discrete algorithm.

By Lemma~\ref{thm:dual_gap_convergence}, the limiting continuous dynamics
converges to the set $\mathcal{N}$ of Nash equilibria. Hence, the only
internally chain-transitive invariant set of the limiting dynamics is
contained in $\mathcal{N}$. \cite{benaim2005}'s stochastic approximation result then
implies that the limit set of the discrete iterates is almost surely
contained in $\mathcal{N}$. Therefore,
\[
    \operatorname{dist}
    \left(
        \Policy{}{}^k,
        \mathcal{N}
    \right)
    \xrightarrow[k\to\infty]{\mathrm{a.s.}}
    0.
\]
Thus, the algorithm converges almost surely to the set of Nash
equilibria.
\end{proof}

\section{Extensive Form Game}
\label{extensive_form_game}

\subsection{A Hierarchy of Planners}

\cite{Escudie2025} introduce a hierarchy of planners for zero-sum partially observable stochastic games (zs-POSGs), based on the observation that different planning procedures arise from progressively stronger assumptions about how future decisions are treated during planning. In particular, the hierarchy distinguishes planners according to the information they have access to and the extent to which future decisions—both their own and those of the opponent—are treated as fixed or optimized during planning.

At the two levels relevant here, the hierarchy can be summarized as follows:
\begin{itemize}
\item \textbf{Focal planner:} a player-specific planning process in which only the focal player's policy is known. The planner therefore reasons about its own decisions without access to the opponent's policy.

\item \textbf{Uninformed planner:} a planner that additionally knows the opponent's policy, together with the information available to the focal planner. This setting can be formulated as a transition-independent zero-sum stochastic game, and is closely related to the stochastic-game formulation introduced at the beginning of this paper.
\end{itemize}

Beyond this operational characterization, we argue that the hierarchy admits a deeper structural interpretation. To make this interpretation explicit, we first consider the example of \cite{SokotaDL0KB23}, which illustrates why a reduction based solely on public policies can fail. In particular, their discussion of the \emph{Backward Dependence Problem} highlights how future decisions may depend on information and decisions made earlier in the game, even when these dependencies are not captured by the public policy representation alone.

\subsection{Equilibrium and Dynamic Programming}
The distinction between simultaneous- and sequential-move games provides a deeper motivation for the hierarchy of planners introduced by \cite{Escudie2025}. We illustrate this point by revisiting the example of \cite{SokotaDL0KB23} and comparing it with the approach developed in \cite{Escudie2025}.

In a simultaneous-move POSG, the equilibrium sought by the dynamic-programming formulation is a Subgame Perfect Nash Equilibrium (SPNE). This is reflected directly in the corresponding Bellman recursion. In the two-player zero-sum case, the backward recursion can be written schematically as

$$
    NE_{(\Policy{1}{0}, \Policy{2}{0})}
    NE_{(\Policy{1}{1}, \Policy{2}{1})}
    \cdots
    NE_{(\Policy{1}{\Horizon - 1}, \Policy{2}{\Horizon - 1})}
    v^{\Policy{1}{}, \Policy{2}{}},
$$

where $NE$ denotes a Nash equilibrium of the corresponding subgame. Thus, the equilibrium is constructed recursively, with the decision rules at each stage being optimized within the corresponding subgame. Because the recursion solves each subgame, the resulting strategy profile satisfies the subgame-perfectness requirement.

This correspondence between dynamic programming and SPNE does not hold in the same way for sequential POSGs, and consequently for extensive-form games (EFGs). In an EFG, the Nash equilibrium sought by the game formulation need not be subgame perfect, even after a reduction to public policies.

This distinction is particularly apparent in Kuhn Poker, where the optimal decision rule at the final stage is not deterministic. If the Bellman recursion above is nevertheless applied directly to the sequential game after reducing it to a public-policy representation, the final stage is optimized through a $\max$ or $\min$ over the available decision rules. This optimization selects a deterministic decision rule at the final stage and therefore fails to capture the dependence of the final decision on deviations at earlier stages. \cite{SokotaDL0KB23} identify the same structural issue in their example of the ``Backward Dependence Problem.''

This observation motivates changing the equilibrium formulation rather than simply applying the simultaneous-game Bellman recursion to a sequential game. Nevertheless, an extensive-form game can be transformed into a lossless game for which a Subgame Perfect Nash Equilibrium exists. The resulting formulation is closely related to the reduction introduced at the beginning of this paper.

\subsection{One-Sided Extensive-Form Games}

How should we handle terminal values in a finite-horizon formulation? In an extensive-form game (EFG), terminal nodes may be reached before the last possible time step, whereas the dynamic-programming formulation evaluates the value at the final time step. Without loss of generality, we can postpone the realization of any terminal reward until the final time step by introducing absorbing states. More precisely, whenever a terminal node is reached at time $t < \Horizon-1$, we transition to an absorbing state and keep the process there until time $\Horizon-1$, at which point the corresponding terminal reward is received. This transformation preserves the value of every policy profile while ensuring that all rewards are realized at the final time step. This is done for readability and to maintain consistency across the different game formulations, thereby simplifying the notation and equations.

\begin{definition}
\label{def:one_sided_extensive_form_game}
Let $\Player{i}$ be a player.We define the associated \textbf{one-sided EFG (oEFG)} as

$$
\EFG{i} \doteq (\Agents, \mathcal{D}, \mathbf{H}, \mathbf{H}^k, \mathcal{D}^i_t, u),
$$

where the components are defined as follows:
\begin{itemize}
\item $\Agents{}$ is the set of $n$ agents in the original game.

\item $\mathcal{D}$ is the decision tree. Each node at time $t$ represents a joint policy prefix
\[
    \Policy{}{0:t}
    =
    \{\Policy{1}{0}, \Policy{2}{0}, \ldots, \Policy{k}{t}\},
\]
where $k$ denotes the player $\Player{k}$. A child node $\Policy{}{}'$ is accessible from $\Policy{}{}$ if
\[
    \exists\, \Policy{k}{t}
    \quad\text{such that}\quad
    \Policy{}{}'
    =
    [\Policy{}{},\Policy{k}{t}].
\]

\item $\mathbf{H}^k$ is the information partition of player $k$:
\[
\mathbf{H}^k(\Policy{}{0:t}) \doteq
\begin{cases}
    \Policy{k}{0:t} & \text{if } k=i,\\
    \Policy{}{0:t} & \text{otherwise.}
\end{cases}
\]
Thus, the focal player $\Player{i}$ observes only her own policy prefix, whereas the opponent $\Player{-i}$ observes the complete joint policy prefix.

\item $\mathcal{D}^k_t = \PolicySpace{k}{t}$ is the space of decision rules available to player $k$ at time $t$.

\item $u$ is the reward function defined on terminal nodes:
\[
    u(\Policy{}{0:\Horizon})
    \doteq
    \ValueFunction{}{}^{\Policy{1}{0:\Horizon},\Policy{2}{0:\Horizon}}.
\]

\end{itemize}

The players make their decisions sequentially, with $\Player{i}$ choosing first and $\Player{-i}$ choosing after observing the complete policy prefix of $\Player{i}$. Consequently, the value of the game is characterized by a max-min optimization: $\Player{i}$ chooses a decision rule to maximize the value, while $\Player{-i}$ observes this choice and subsequently chooses a decision rule to minimize it.
\end{definition}

Under this construction, $\Player{i}$ has access only to her own policy, whereas $\Player{-i}$ has access to both players' policies. We then show that any zero-sum EFG admits a lossless reduction to the corresponding zero-sum oEFG.

\begin{theorem}
Let $\EFG{}$ be a zero-sum extensive-form game and let $\Player{i}$ be the focal player. The reduction

$$
    \mathcal{T}_{\EFG{} \rightarrow \EFG{i}}
$$

that projects $\EFG{}$ onto the one-sided EFG $\EFG{i}$ is \textit{lossless} for player $\Player{i}$.
In particular, it preserves the value of every joint policy and the focal player's maximin value and optimal policies.
\end{theorem}

\begin{proof}
We establish the result by showing that the reduction preserves (i) the correspondence between joint policies and paths in the decision tree, (ii) the associated values, and (iii) the information available to the focal player.

\textbf{Policy and path correspondence.}
Consider a joint policy

$$
    \Policy{}{0:\Horizon}
    =
    \bigl(
        \Policy{i}{0:\Horizon},
        \Policy{-i}{0:\Horizon}
    \bigr)
$$

in the original EFG $\EFG{}$. The reduction associates with this joint policy a policy $\Tilde{\Policy{}{}}$ in the one-sided EFG $\EFG{i}$.

At each node corresponding to a policy prefix $\Policy{}{0:t}$, the next decision of the focal player is determined by her own policy prefix $\Policy{i}{0:t}$, whereas the next decision of the opponent is determined by the complete policy prefix $\Policy{}{0:t}$. Thus, the complete joint policy uniquely determines a path in the decision tree $\mathcal{D}$, and conversely, a path in $\mathcal{D}$ specifies the sequence of decision rules constituting the corresponding joint policy.

\textbf{Value preservation.}
By construction, the terminal node reached by the path associated with
$\Policy{}{0:\Horizon}$ has reward

$$
    u(\Policy{}{0:\Horizon})
    \doteq
    v^{\Policy{i}{0:\Horizon},\Policy{-i}{0:\Horizon}}.
$$

Consequently, the expected value induced by any joint policy is unchanged by the reduction:

$$
    u(\Tilde{\Policy{}{}})
    =
    v^{\Policy{i}{0:\Horizon},\Policy{-i}{0:\Horizon}}.
$$

Hence, optimizing over corresponding policies in the two representations yields the same payoff. In particular, the reduction does not introduce or remove any payoff-relevant trajectories.

\textbf{Information preservation for the focal player.}
The reduction is one-sided precisely because it preserves the information available to the focal player while explicitly exposing additional information to the opponent. In $\EFG{i}$, player $\Player{i}$ observes only her own policy prefix through the information partition
$\mathbf{H}^{i}$:

$$
    \mathbf{H}^{i}(\Policy{}{0:t})
    =
    \Policy{i}{0:t}.
$$

This is exactly the information on which her decision rule is allowed to depend. By contrast, the opponent observes the complete policy prefix,

$$
    \mathbf{H}^{-i}(\Policy{}{0:t})
    =
    \Policy{}{0:t},
$$

and can therefore condition its decision on the focal player's previous decisions. The opponent thus acts as an informed best responder. Since the theorem concerns losslessness for player $\Player{i}$, granting this additional information to the opponent does not alter the focal player's policy space or information structure; rather, it makes the opponent's minimization explicit.

\textbf{Preservation of the focal player's maximin problem.}
Because the game is zero-sum, the value associated with a joint policy can be written as

$$
    v^{\Policy{i}{},\Policy{-i}{}}.
$$

The focal player's security problem in the original game is therefore

$$
    \max_{\Policy{i}{}}
    \min_{\Policy{-i}{}}
    v^{\Policy{i}{},\Policy{-i}{}}.
$$

The policy and value correspondence established above implies that the same optimization problem is represented in $\EFG{i}$:

\[
    \max_{\tilde{\Policy{i}{}}}\min_{\tilde{\Policy{-i}{}}}u(\Tilde{\Policy{}{}}).
\]

Thus, the two formulations have the same maximin value, and a policy is optimal for player $\Player{i}$ in the original EFG if and only if its corresponding policy is optimal in the one-sided EFG.

Finally, by the minimax theorem for finite zero-sum games, the maximin value coincides with the value attained at a Nash equilibrium. Hence, the focal player's equilibrium value and the set of focal-player maximin strategies are preserved by the reduction, even though the one-sided representation does not claim to preserve the opponent's information structure or the full set of equilibrium strategies.
\end{proof}

\tikzset
{
    rev_label/.style={anchor=south, rotate=-90, inner sep=2mm}
}

\tikzset
{
    hor_label/.style={anchor=south, inner sep=2mm}
}

\tikzset
{
    diag_label/.style={anchor=south, rotate=-35, inner sep=2mm}
}

\begin{figure}[h]
    \begin{center}
        \begin{tikzcd}[
            row sep=7em, 
            column sep=7em,    
            shorten >=10pt,  
            shorten <=10pt   
        ]
            |[
                fill=white,
                inner sep=4pt,
                label={[yshift=0.4em, font=\scriptsize\itshape, fill=white, inner sep=2pt]below:\text{Nash Equilibrium}}
            ]|\text{zs-POSG $\GameTuple$} 
                \arrow[r, "\mathcal{T}_{POSG \rightarrow EFG}" hor_label] 
                \arrow[d] 
                \arrow[rd, "\mathcal{T}_{POSG \rightarrow EFG} \circ \mathcal{T}_{EFG \rightarrow OEFG}" diag_label] 
            & |[
                fill=white,
                inner sep=4pt,
                label={[yshift=0.4em, font=\scriptsize\itshape, fill=white, inner sep=2pt]below:\text{Nash Equilibrium}}
            ]|\text{zs-EFG $\EFG{}$} 
                \arrow[d, "\mathcal{T}_{EFG \rightarrow OEFG}" rev_label] \\
            |[
                fill=white,
                inner sep=4pt,
                label={[yshift=0.4em, font=\scriptsize\itshape, fill=white, inner sep=2pt]below:\text{Subgame Perfect NE}}
            ]|\text{zs-SG $\GameTuple'$}  
                \arrow[d]
            & |[
                fill=white,
                inner sep=4pt,
                label={[yshift=0.4em, font=\scriptsize\itshape, fill=white, inner sep=2pt]below:\text{Subgame Perfect NE}}
            ]|\text{zs-OEFG $\EFG{i}$} \\
            |[
                fill=white,
                inner sep=4pt,
                label={[yshift=0.4em, font=\scriptsize\itshape, fill=white, inner sep=2pt]below:\text{Subgame Perfect NE}}
            ]|\text{zs-OSG $\FocalBeliefGameTuple{i}$} 
        \end{tikzcd}
    \end{center}
    \caption{$\mathcal{T}_{\mathcal{G} \rightarrow \mathcal{G}'}$ denotes the reduction from game $\mathcal{G}$ to game $\mathcal{G}'$. The reduction $\mathcal{T}_{POSG \rightarrow EFG}$ is almost transparent: it simply reformulates the POSG as an extensive-form game (EFG). However, unlike $\mathcal{T}_{POSG \rightarrow SG}$, this reduction does not induce an equilibrium concept that can be characterized inductively in the reduced game. The composition $\mathcal{T}_{POSG \rightarrow EFG} \circ \mathcal{T}_{EFG \rightarrow OEFG}$ corresponds to the focal-planner reduction introduced in \cite{Escudie2025}.
}
    \label{fig:EFG_reduction}
\end{figure}
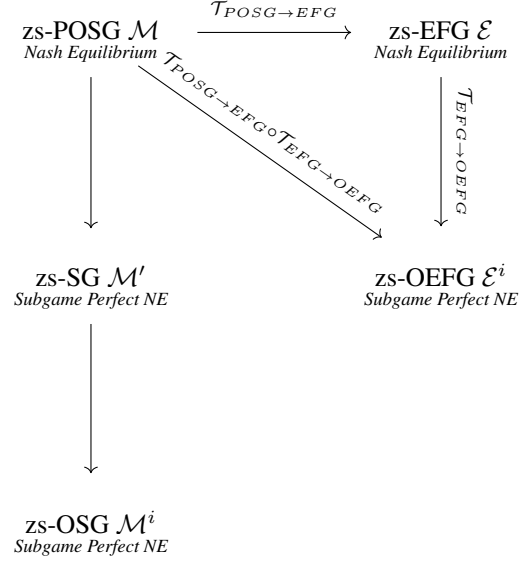

\begin{theorem}
\label{lem:efg_lossless_reduction}
Under the convention introduced in Definition~\ref{def:one_sided_extensive_form_game}, the one-sided EFG $\EFG{i}$ admits a unilateral Subgame Perfect Nash Equilibrium. In particular, its value satisfies the following Bellman recursion:

$$
    v^{*,i}_{t}(\Policy{i}{0:t-1})
    =
    \max_{\Policy{i}{t}}
    \min_{\Policy{-i}{0:\Horizon-1}}
    v^{*,i}_{t+1}(\Policy{i}{0:t}),
$$

where $v^{*}_{t}(\Policy{i}{0:t-1})$ denotes the optimal continuation value for player $\Player{i}$ after fixing her policy prefix $\Policy{i}{0:t-1}$.

At the final decision stage, the recursion reduces to

$$
    v^{*,i}_{\Horizon-1}(\Policy{i}{0:\Horizon-2})
    =
    \max_{\Policy{i}{\Horizon-1}}
    \min_{\Policy{-i}{0:\Horizon-1}}
    u_{\Horizon-1}
    \left(
        \Policy{i}{0:\Horizon-1},
        \Policy{-i}{0:\Horizon-1}
    \right).
$$

\end{theorem}

\begin{proof}
We prove by backward induction the stronger statement $(*)$,

$$
\begin{aligned}
    v^{*,i}_{t}(\Policy{i}{0:t-1})
    =
    \max_{\Policy{i}{t:\Horizon-1}}
    \min_{\Policy{-i}{0:\Horizon-1}}
    u\left(
        [\Policy{i}{0:t-1},\Policy{i}{t:\Horizon-1}],
        \Policy{-i}{0:\Horizon-1}
    \right)
\end{aligned}
$$
for every $t$.

\textbf{Base case.}
At $t=\Horizon-1$, there is no future decision of player $\Player{i}$ after $\Policy{i}{\Horizon-1}$. Therefore, by definition,

$$
    v^{*,i}_{\Horizon-1}(\Policy{i}{0:\Horizon-2})
    =
    \max_{\Policy{i}{\Horizon-1}}
    \min_{\Policy{-i}{0:\Horizon-1}}
    u\left(
        [\Policy{i}{0:\Horizon-2},\Policy{i}{\Horizon-1}],
        \Policy{-i}{0:\Horizon-1}
    \right),
$$

so property $(*)$ holds.

\textbf{Induction step.}
Assume that $(*)$ holds at time $t+1$. Then
\begin{align*}
v^{*,i}_{t}(\Policy{i}{0:t-1})
&=
\max_{\Policy{i}{t}}
\min_{\Policy{-i}{0:\Horizon-1}}
v^{*,i}_{t+1}(\Policy{i}{0:t})\\
&=
\max_{\Policy{i}{t}}
\min_{\Policy{-i}{0:\Horizon-1}}
\max_{\Policy{i}{t+1:\Horizon-1}}
\min_{\Policy{-i}{0:\Horizon-1}}
u\left(
[\Policy{i}{0:t-1},
\Policy{i}{t},
\Policy{i}{t+1:\Horizon-1}],
\Policy{-i}{0:\Horizon-1}
\right).
\end{align*}

We now justify the exchange of the middle $\min$ and $\max$. Fix a prefix $\Policy{i}{0:t}$. Player $\Player{i}$ chooses a complete continuation policy

$$
    \Policy{i}{t+1:\Horizon-1},
$$

while player $\Player{-i}$ chooses a complete policy

$$
    \Policy{-i}{0:\Horizon-1}.
$$

For fixed $\Policy{i}{0:t}$, let the payoff of the corresponding pure-strategy pair be

$$
    U\left(
        \Policy{i}{t+1:\Horizon-1},
        \Policy{-i}{0:\Horizon-1}
    \right)
    =
    u\left(
        [\Policy{i}{0:t},
        \Policy{i}{t+1:\Horizon-1}],
        \Policy{-i}{0:\Horizon-1}
    \right).
$$

When the two players randomize over these complete policies, the expected payoff is bilinear in their mixed strategies. Hence, by the minimax theorem,

$$
    \min_{\Policy{-i}{0:\Horizon-1}}
    \max_{\Policy{i}{t+1:\Horizon-1}}
    U
    =
    \max_{\Policy{i}{t+1:\Horizon-1}}
    \min_{\Policy{-i}{0:\Horizon-1}}
    U.
$$

This is the same minimax argument used in \cite{DelBufDibSaf-DGAA-23}. Importantly, the argument is applied only after fixing the focal player's prefix $\Policy{i}{0:t}$; the continuation policies of the two players then constitute the strategy spaces of the induced zero-sum normal-form game.

Applying this equality inside the backward recursion gives
\begin{align*}
v^{*,i}_{t}(\Policy{i}{0:t-1})
&=
\max_{\Policy{i}{t}}
\max_{\Policy{i}{t+1:\Horizon-1}}
\min_{\Policy{-i}{0:\Horizon-1}}
u\left(
[\Policy{i}{0:t-1},
\Policy{i}{t},
\Policy{i}{t+1:\Horizon-1}],
\Policy{-i}{0:\Horizon-1}
\right)\\
&=
\max_{\Policy{i}{t:\Horizon-1}}
\min_{\Policy{-i}{0:\Horizon-1}}
u\left(
[\Policy{i}{0:t-1},\Policy{i}{t:\Horizon-1}],
\Policy{-i}{0:\Horizon-1}
\right).
\end{align*}
This is precisely property $(*)$ at time $t$, completing the induction.

Finally, applying $(*)$ at $t=0$ yields

$$
    v^{*,i}_{0}
    =
    \max_{\Policy{i}{0:\Horizon-1}}
    \min_{\Policy{-i}{0:\Horizon-1}}
    u\left(
        \Policy{i}{0:\Horizon-1},
        \Policy{-i}{0:\Horizon-1}
    \right).
$$

Thus, the policy generated by the Bellman recursion is a maximin-optimal policy for player $\Player{i}$. Since the opponent observes the complete policy prefix of player $\Player{i}$, this sequential maximin construction corresponds to the unilateral Subgame Perfect Nash Equilibrium of the one-sided EFG. Hence, the Bellman recursion computes the equilibrium value and an equilibrium strategy for the focal player.
\end{proof}

A few observations follow from this construction. First, by reducing an EFG to a one-sided EFG, we obtain a well-defined notion of a proper subgame for the sequential game. This allows us to formulate and compute a Subgame Perfect Nash Equilibrium, thereby addressing the backward-dependence issue identified by \cite{SokotaDL0KB23}. In particular, the one-sided construction makes the information available to the two players explicit and ensures that the continuation problem at each stage is formulated as the appropriate sequential zero-sum game.

Second, this result is closely related to Theorem~3 of \cite{Escudie2025}. While our proof is formulated directly in terms of the decision tree and does not explicitly use the occupancy-state representation, the correspondence between the two formulations is immediate: the nodes of the one-sided EFG can equivalently be represented by the corresponding occupancy states. Thus, the construction provides an alternative, more explicit interpretation of the same underlying reduction.

The main distinction is the scope in which this construction is required. The one-sided reduction is necessary when dealing with sequential games, such as EFGs or sequential POSGs, because future decisions may depend on previously revealed decisions and therefore cannot in general be optimized independently through the standard simultaneous-move Bellman recursion. In contrast, most of the games considered in this paper are simultaneous-move POSGs. In that setting, the standard dynamic-programming formulation already corresponds to the appropriate subgame-perfect solution concept, and the additional one-sided reduction is therefore unnecessary. The relationship between these two cases is illustrated in Figure~\ref{fig:EFG_reduction}.

We work with the uninformed planner introduced by \cite{Escudie2025} because deviations in the past cannot depend on the public history. If we were to use an informed planner instead, the resulting optimization would allow a different deviation for each public history. This would effectively give the best response access to the public history before the corresponding decision is made, which is not available to the player and would therefore be inconsistent with the information structure of the game.

\subsection{Experiments}

We extend Algorithm~$\ref{alg:api}$ to extensive-form games by adapting the local policy-improvement step to account for deviations from decisions made at earlier stages of the game. In particular, the linear program is modified to retain the relevant information about past decisions when computing the current deviation, allowing the local improvement procedure to operate directly on the extensive-form representation.

To obtain a deviation model consistent with the intended information structure, we include public histories in the support of the belief representation. The deviation is therefore defined with respect to the information available to the player rather than separately for each public history. Without this restriction, the optimization could effectively select a different deviation for every public history, leading to an overly pessimistic evaluation that does not correspond to the intended player model.

We implemented this extension and evaluated it on Kuhn Poker. The resulting algorithm exhibits convergence to the expected equilibrium behavior in our experiments. However, the current implementation is primarily intended as a proof-of-concept: the code requires further cleanup and validation before the experimental results can be presented as part of the main empirical evaluation. We therefore leave a more systematic experimental study of the extensive-form extension to future work.




\section{Experiments}

\subsection{Baselines}

\paragraph{PBVI in zs-SG} In a POMDP, the value function is defined over the belief simplex and is piecewise-linear and convex, making exact value iteration intractable. PBVI addresses this issue by restricting computation to a finite set of representative beliefs. At each iteration, Bellman backups are performed only on this set, yielding an approximate value function that is accurate in the vicinity of the sampled beliefs while remaining tractable.

PBVI is inherently a single-agent method. In adversarial settings, it is therefore applied to a one-sided (focal-planner) surrogate, where one player is optimized while the opponent is treated as part of the environment. Two levels of sampling arise. First, beliefs are expanded via forward simulation to cover reachable regions of the belief space. Second, in the one-sided reduction, the opponent is simulated (e.g., via randomized policies), so that the focal player optimizes against sampled behaviors. This relies on the assumption that relevant opponent responses lie near those induced by sampling.

Each PBVI backup requires a maximization over actions and observations, which in adversarial settings leads to linear programs for worst-case evaluation. The number of $\alpha$-vectors grows rapidly with iterations, increasing the size of these programs. Consequently, the Bellman operator quickly becomes intractable despite the point-based approximation.

\paragraph{CFR in POSG.}
\label{appendix:ocfr}
Counterfactual Regret Minimization (CFR) is an iterative algorithm for solving extensive-form games (EFGs) that minimizes counterfactual regret at each information set, yielding convergence to a Nash equilibrium in two-player zero-sum settings. At each iteration, CFR evaluates the current strategy profile, computes counterfactual values, and updates strategies via regret matching.

The connection between POSGs and EFGs has been established in \citep{Kovarik2022}, laying the groundwork for the application of extensive-form techniques such as CFR. Notably, the theoretical objects introduced in our paper naturally embed CFR and yield a refined variant tailored to POSGs. To make this connection explicit, Table~\ref{tab:efg-to-posg} presents a correspondence between standard EFG notions and their POSG counterparts, thereby providing a bridge for transferring the algorithm. The relevant notation is summarized in Table~\ref{tab:efg-to-posg}.

\begin{table}[t]
\centering
\small
\setlength{\tabcolsep}{15pt}
\renewcommand{\arraystretch}{1.}
\begin{tabular}{p{2.8cm} c l c}
\toprule
\textbf{Name in EFG} & \textbf{Symbol} & \textbf{Name in POSG} & \textbf{Symbol} \\

\toprule

History & $h$ & State and Joint history & $(s_t, h_t)$\\


i's information set & $I$ & i's private history & $h_i$\\


Strategy profile & $\sigma$ & Joint policy & $\pi$\\


i's private strategy & $\sigma_{i}$ & i's private policy & $\pi_{i}$\\


Players' contribution except i & $\pi^{\sigma}_{i}(I)$ & Players' contribution except i & $\Pr(h_i \mid \pi, b_0)$\\


Players' contribution except i & $\pi^{\sigma}_{i}(h)$ & Counterfactual state & $c_{i}(s, h_{-i})$\\


Counterfactual value of $\sigma$ at $h$ & $v_{i}(\sigma,h)$ & Value of $\pi$ at $(s,h)$ & $v^{\pi}_{\GameTuple}(s, h)$\\


Deviation at $I$ & $\sigma_{I \rightarrow a}$ & Deviation at $h_{i}$ & $\pi_{h_{i} \rightarrow a_{i}}$\\

\bottomrule

\end{tabular}
\caption{Translation from EFG to POSG. For more information, we refer the reader to \citep{Kovarik2022}. We connect the history $h$ in EFG to the pair $(s,h)$ in POSG since the reward is not associated to the history but to the state in POSG. The quantity $v^{\pi}_{\GameTuple}(s, h)$ is formally defined in \citep{cunha2023convex}, we refer the reader for more information. It is simply the expected cumulated value a player can get at this pair.}
\label{tab:efg-to-posg}
\end{table}








From \citep{Zinkevich2007,Neller2013AnIT}, we adapt the key ingredients of CFR to the POSG setting. We first define the (instantaneous) regret at $(s_t,h_t)$ for player $-i$ when deviating only at $h_{-i,t}$ to action $a_{-i,t}$:
\[
r_{\GameTuple}(s_t,h_t,a_{i,t})
= v^{\pi_{h_{i,t} \rightarrow a_{i,t}}}_{\GameTuple}(s_t,h_t)
- v^{\pi}_{\GameTuple}(s_t,h_t).
\]
In contrast to EFGs, where regret is indexed by histories $h$, in POSGs it is defined over $(s_t,h_t)$ since rewards depend on the (hidden) state $s_t$. The history remains necessary to condition the belief over states.

Aggregating over histories that share the same private history $h_{i,t}$ yields a vector-valued regret:
\[
\mathbf{r}^{h_{i,t} \rightarrow a_{i,t}}_{\GameTuple}(s_t,h_{i,t})
= v^{\pi_{h_{-i,t} \rightarrow a_{-i,t}}}_{\GameTuple}(s_t,h_{i,t},h_{-i,t})
- v^{\pi}_{\GameTuple}(s_t,h_{i,t},h_{-i,t}),
\]
which is a vector over $\mathcal{C}_{i}$ (one coordinate per state). Using the conditional state $c_i$ notation, we recover a scalar regret:
\[
R^{h_{-i,t} \rightarrow a_{-i,t}}_{\GameTuple}
= \mathbf{r}^{h_{-i,t} \rightarrow a_{-i,t}} \cdot c_i.
\]
This corresponds to the EFG quantity
\[
r_{EFG}(I,a)
= \sum_{h \in I} \bigl(v_{i}(\sigma_{I \rightarrow a},h) - v_{i}(\sigma,h)\bigr),
\]
where the summation over histories is replaced by expectation under the counterfactual occupancy.

The policy is then updated via regret matching:
\[
\pi_{-i,\text{new}}(a_{-i,t} \mid h_{-i,t})
= \frac{\bigl[R^{h_{-i,t} \rightarrow a_{-i,t}}_{\GameTuple}\bigr]_+}
{\sum_{a'_{-i,t}} \bigl[R^{h_{-i,t} \rightarrow a'_{-i,t}}_{\GameTuple}\bigr]_+}.
\]

The regret of player $-i$ is inherently \emph{vector-valued}, as it is defined conditionally on the hidden state. The conditional state $c_i$ acts as a belief, projecting this vector onto a scalar quantity, exactly as counterfactual reach probabilities do in EFGs. This establishes a direct connection with the focal-planner view: regret minimization operates on belief-weighted values, consistent with optimizing a surrogate game for a fixed opponent.

Our framework can be interpreted as performing regret minimization directly in the occupancy space. Instead of explicitly maintaining regrets per history, we operate on regret functions over counterfactual occupancy states, and policy updates correspond to regret matching under these occupancy-induced weights. This yields a compact implementation that avoids explicit enumeration of histories.





\definecolor{verylightgray}{gray}{0.9}

\begin{table}
    \setlength\extrarowheight{-3pt}

    \caption{Snapshot of empirical results. For each game, horizon and algorithm, we report time, value and exploitability $\epsilon$ associated with the computed $\epsilon$-Nash equilibrium. {\sc OOT} occurs when the target exploitability threshold cannot be reached within the prescribed time limit, while {\sc oom} means that the algorithms ran out of memory.  \textcolor{magenta}{\textbf{magenta}} = lowest exploitability value per benchmark, while `$-$' means the time budget for computing the exploitability has been exceeded.}
    \label{table:main_results:bigtable:exploitability_cost}
    \resizebox{\linewidth}{!}{
        \begin{tabular}{@{}c c rr@{\hspace{4pt}}l rr@{\hspace{4pt}}l rr@{\hspace{4pt}}l rr@{\hspace{4pt}}l}
        
        \toprule%
         \centering%
         & 
         & \multicolumn{3}{c}{{{\bfseries \texttt{CPI} (Ours)}}}
         & \multicolumn{3}{c}{{{\bfseries \texttt{PBVI}}}}
         & \multicolumn{3}{c}{{{\bfseries \texttt{HSVI}}}}
         & \multicolumn{3}{c}{{{\bfseries \texttt{CFR+}}}}
         \\

         \multirow{-4}{*}{\textbf{Problem}} & \multirow{-4}{*}{$T$} & t & val. & $\epsilon$ & t & val. & $\epsilon$ & t & val. & $\epsilon$ & t & val. & $\epsilon$\\ 
        
        \cmidrule[0.4pt](lr{1em}){3-5}%
        \cmidrule[0.4pt](lr{1em}){6-8}%
        \cmidrule[0.4pt](lr{1em}){9-11}%
        \cmidrule[0.4pt](lr{1em}){12-14}%

        \arrayrulecolor{verylightgray}
        
            & 2 & {0.3} & \opti{0.078} & \opti{0.} & \opti{0.2} & \opti{0.078} & \opti{0.} & {0.4} & \opti{0.078} & \opti{0.} & {0.3} & \opti{0.078} & \opti{0.}\\

            
            & 3 & {11} & \opti{0.097} & \opti{0.} & \opti{24} & {0.096} & \opti{0.} & {70} & \opti{0.097} & \opti{0.} & \opti{7} & {0.095} & \opti{0.}\\
            
            
            & 4 & \opti{23} & \opti{0.11} & \opti{0.} & {65} & \opti{0.11} & {0.01} & \textsc{oot} & \textsc{oot} & \textsc{oot} & {25} & \opti{0.11} & {0.01}\\
            
            
            & 5 & \opti{29} & \opti{0.12} & \opti{0.} & {31} & \opti{0.12} & \opti{0.} & \textsc{oot} & \textsc{oot} & \textsc{oot} & {69} & {0.12} & {0.01}\\
            
            
            & 7 & \opti{97} & \opti{0.14} & {0.01} & {190} & \opti{0.14} & \opti{0.0} & \textsc{oot} & \textsc{oot} & \textsc{oot} & {92} & \opti{0.14} & {0.02}\\
            
            
            & 10 & \opti{100} & \opti{0.18} & \opti{0.01} & {637} & {0.17} & \opti{0.01} & \textsc{oot} & \textsc{oot} & \textsc{oot} & {122} & \opti{0.18} & {0.02}\\
            
            
            & 15 & \opti{143} & \opti{0.22} & \opti{0.01} & {668} & {0.22} & {0.02} & \textsc{oot} & \textsc{oot} & \textsc{oot} & {222} & \opti{0.22} & {0.03}\\
            
            
            \multirow{-11}{*}{\textsc{Mabc}} & 30 & \opti{360} & \opti{0.36} & \opti{0.01} & \textsc{oot} & \textsc{oot} & \textsc{oot} & \textsc{oot} & \textsc{oot} & \textsc{oot} & \opti{433} & \opti{0.36} & {0.06}\\
        
        \noalign{\vskip 2pt}
        \arrayrulecolor{black}
        \hline
        \noalign{\vskip 4pt}

            & 2 & \opti{0.01} & \opti{0.4} & \opti{0.} & {2} & {0.39} & \opti{0.} & {0.1} & \opti{0.4} & \opti{0.} & {0.06} & \opti{0.4} & {0.005}\\
            
            & 3 & \opti{0.02} & \opti{0.6} & \opti{0.} & {18} & \opti{0.6} & \opti{0.} & {1} & \opti{0.6} & \opti{0.} & {0.1} & \opti{0.6} & {0.001}\\
            
            & 4 & \opti{0.04} & \opti{0.8} & \opti{0.} & {34} & {0.77} & {0.} & {4} & \opti{0.8} & \opti{0.} & {0.2} & \opti{0.8} & \opti{0.}\\
            
            & 5 & \opti{0.04} & \opti{1.} & \opti{0.} & {52} & {0.95} & \opti{0.} & {60} & \opti{1.} & \opti{0.} & {0.25} & \opti{1.} & \opti{0.}\\
            
            & 7 & \opti{13} & \opti{1.4} & \opti{0.} & {180} & {1.31} & {0.03} & \textsc{oot} & \textsc{oot} & \textsc{oot} & {15} & {1.39} & {0.01}\\
            
            & 10 & \opti{92} & \opti{2.} & \opti{0.} & {\textsc{oot}} & \textsc{oot} & \textsc{oot} & \textsc{oot} & \textsc{oot} & \textsc{oot} & {103} & {1.99} & {0.02}\\

            & 25 & \opti{15} & \opti{5.} & \opti{0.} & \textsc{oot} & \textsc{oot} & \textsc{oot} & \textsc{oot} & \textsc{oot} & \textsc{oot} & {132} & \opti{5.} & {0.02}\\
            
            \multirow{-9}{*}{\shortstack[C]{\textsc{Matching} \\ \textsc{Pennies}}} & 50 & \opti{217} & \opti{10.} & \opti{0.} & \textsc{oot} & \textsc{oot} & \textsc{oot} & \textsc{oot} & \textsc{oot} & \textsc{oot} & {718} & {10.} & {0.03}\\

        \noalign{\vskip 2pt}
        \arrayrulecolor{black}
        \hline
        \noalign{\vskip 4pt}

            & 2 & \opti{0.4} & \opti{0.26} & \opti{0.} & \opti{0.4} & \opti{0.26} & \opti{0.} & {1.2} & \opti{0.26} & \opti{0.} & {2.8} & \opti{0.26} & {0.004}\\
            
            & 3 & \opti{17} & \opti{0.32} & \opti{0.01} & {78} & \opti{0.32} & \opti{0.} & {120} & \opti{0.32} & \opti{0.} & {75} & \opti{0.32} & {0.01}\\
            
            & 4 & \opti{42} & \opti{0.36} & \opti{0.03} & {56} & {0.35} & {0.01} & \textsc{oot} & \textsc{oot} & \textsc{oot} & {104} & {0.35} & {0.04}\\
            
            & 5 & \opti{213} & \opti{0.4} & \opti{0.05} & {98} & {0.38} & {0.05} & \textsc{oot} & \textsc{oot} & \textsc{oot} & {177} & {0.38} & {0.05}\\
            
            & 7 & \opti{271} & \opti{0.5} & \opti{0.06} & {1481} & {0.48} & {0.04} & \textsc{oot} & \textsc{oot} & \textsc{oot} & {503} & \opti{0.49} & {0.12}\\

            & 10 & \opti{505} & {0.63} & \opti{0.07} & {3232} & {0.6} & \opti{0.06} & \textsc{oot} & \textsc{oot} & \textsc{oot} & {1151} & \opti{0.64} & {0.15}\\
            
            \multirow{-9}{*}{\textsc{Recycling}} & 15 & \opti{703} & \opti{0.87} & \opti{0.08} & \textsc{oot} & \textsc{oot} & \textsc{oot} & \textsc{oot} & \textsc{oot} & \textsc{oot} & {2415} & {0.86} & {0.23}\\

        \noalign{\vskip 2pt}
        \arrayrulecolor{black}
        \hline
        \noalign{\vskip 4pt}
        
            & 2 & \opti{0.1} & \opti{-0.4} & \opti{0.} & \opti{0.1} & \opti{-0.4} & \opti{0.} & {0.2} & \opti{-0.4} & \opti{0.} & \opti{0.1} & \opti{-0.4} & \opti{0.0}\\
            
            & 3 & \opti{0.2} & \opti{-0.56} & \opti{0.} & {0.3} & \opti{-0.56} & \opti{0.} & {500} & \opti{-0.56} & \opti{0.} & {5} & \opti{-0.56} & \opti{0.}\\
            
            & 4 & \opti{16} & \opti{-0.75} & {0.01} & {46} & {-0.76} & \opti{0.02} & \textsc{oot} & \textsc{oot} & \textsc{oot} & {76} & \opti{-0.75} & {0.03}\\

            & 5 & {82} & \opti{-0.96} & {0.05} & {141} & {-0.96} & \opti{0.05} & \textsc{oot} & \textsc{oot} & \textsc{oot} & \opti{87} & \opti{-0.96} & {0.07}\\

            & 7 & \opti{248} & {-1.36} & {0.13} & {866} & {-1.36} & \opti{0.1} & \textsc{oot} & \textsc{oot} & \textsc{oot} & {172} & \opti{-1.35} & {0.13}\\
            
            \multirow{-7}{*}{\shortstack[C]{\textsc{Adversarial} \\ \textsc{Tiger}}} & 10 & {824} & {-1.97} & {0.24} & {956} & {-1.98} & {0.18} & \textsc{oot} & \textsc{oot} & \textsc{oot} & {556} & {-1.95} & {0.3}\\

        \bottomrule
        \end{tabular}
        \vspace{-.5cm}
        }
\end{table}

\begin{table}
    \setlength\extrarowheight{-3pt}
    \caption{A snapshot of empirical results with no pruning mechanism applied (see \cite{Escudie2025}, \cite{Dibangoye:OMDP:2016}). For each game, horizon and algorithm, we report time, value and exploitability $\epsilon$ associated with the computed $\epsilon$-Nash equilibrium. {\sc OOT} occurs when the target exploitability threshold cannot be reached within the prescribed time limit, while {\sc oom} means that the algorithms ran out of memory.  \textcolor{magenta}{\textbf{magenta}} = lowest exploitability value per benchmark, while `$-$' means the time budget for computing the exploitability has been exceeded.}
    \resizebox{\linewidth}{!}{
        \begin{tabular}{@{}c c rr@{\hspace{4pt}}l rr@{\hspace{4pt}}l rr@{\hspace{4pt}}l rr@{\hspace{4pt}}l}
        
        \toprule%
         \centering%
         & 
         & \multicolumn{3}{c}{{{\bfseries \texttt{CPI} (Ours)}}}
         & \multicolumn{3}{c}{{{\bfseries \texttt{PBVI}}}}
         & \multicolumn{3}{c}{{{\bfseries \texttt{HSVI}}}}
         & \multicolumn{3}{c}{{{\bfseries \texttt{CFR+}}}}
         \\

         \multirow{-4}{*}{\textbf{Problem}} & \multirow{-4}{*}{$\Horizon$} & t & val. & $\epsilon$ & t & v & $\epsilon$ & t & val. & $\epsilon$ & t & val. & $\epsilon$\\ 
        
        \cmidrule[0.4pt](lr{1em}){3-5}%
        \cmidrule[0.4pt](lr{1em}){6-8}%
        \cmidrule[0.4pt](lr{1em}){9-11}%
        \cmidrule[0.4pt](lr{1em}){12-14}%
        
            & 2 & {0.3} & \opti{0.078} & \opti{0.} & \opti{0.2} & \opti{0.077} & \opti{0.} & {0.3} & \opti{0.078} & \opti{0.} & {0.3} & \opti{0.078} & {0.003} \\
            
            & 3 & \opti{5} & \opti{0.097} & \opti{0.} & {19} & \opti{0.097} & \opti{0.} & {70} & \opti{0.097} & \opti{0.} & {12} & {0.096} & {0.002} \\
            
            & 4 & \opti{38} & \opti{0.11} & \opti{0.001} & {50} & {0.1} & \opti{0.001} & \textsc{oot} & \textsc{oot} & \textsc{oot} & {73} & \opti{0.11} & {0.003} \\
            
            \multirow{-5}{*}{\textsc{Mabc}} & 5 & {2176} & \opti{0.12} & {0.01} & {1365} & \opti{0.12} & {0.013} & \textsc{oot} & \textsc{oot} & \textsc{oot} & {1614} & \opti{0.12} & \opti{0.005} \\

        \noalign{\vskip 2pt}
        \arrayrulecolor{black}
        \hline
        \noalign{\vskip 4pt}

            & 2 & \opti{0.01} & \opti{0.4} & \opti{0.} & {0.13} & {0.38} & \opti{0.} & {0.1} & {0.4} & {0.} & \opti{0.01} & \opti{0.4} & \opti{0.} \\
            
            & 3 & \opti{0.05} & \opti{0.6} & \opti{0.} & {9} & \opti{0.6} & \opti{0.} & {1} & {0.6} & {0.} & {0.2} & \opti{0.6} & \opti{0.} \\
            
            & 4 & \opti{0.5} & \opti{0.8} & \opti{0.} & {250} & {0.77} & \opti{0.} & {4} & {0.8} & {0.} & {2} & \opti{0.8} & {0.005} \\
            
            \multirow{-5}{*}{\shortstack[C]{\textsc{Matching} \\ \textsc{Pennies}}} & 5 & {11} & \opti{1.0} & \opti{0.003} & {3207} & {1.14} & {0.12} & {60} & {1.} & {0.} & \opti{5} & {1.} & {0.006}\\

        \noalign{\vskip 2pt}
        \arrayrulecolor{black}
        \hline
        \noalign{\vskip 4pt}
        
            & 2 & {0.4} & \opti{0.26} & \opti{0.} & \opti{0.1} & \opti{0.26} & \opti{0.} & {1.2} & \opti{0.26} & \opti{0.} & {1.5} & \opti{0.26} & {0.01} \\
            
            & 3 & \opti{14} & \opti{0.32} & \opti{0.} & {286} & \opti{0.32} & \opti{0.} & {120} & \opti{0.32} & \opti{0.} & {131} & \opti{0.32} & {0.01} \\
            
            & 4 & {681} & \opti{0.35} & \opti{0.} & \opti{418} & {0.34} & \opti{0.01} & \textsc{oot} & \textsc{oot} & \textsc{oot} & {848} & \opti{0.35} & {0.03} \\
            
            \multirow{-5}{*}{\textsc{Recycling}} & 5 & \opti{5146} & \opti{0.36} & \opti{0.05} & \textsc{oot} & \textsc{oot} & \textsc{oot} & \textsc{oot} & \textsc{oot} & \textsc{oot} & {7611} & \opti{0.36} & {0.1}\\

        \noalign{\vskip 2pt}
        \arrayrulecolor{black}
        \hline
        \noalign{\vskip 4pt}
        
            & 2 & \opti{0.} & \opti{-0.4} & \opti{0.} & \opti{0.} & \opti{-0.4} & \opti{0.} & \opti{0.} & \opti{-0.4} & \opti{0.} & \opti{0.} & \opti{-0.4} & {0.002} \\
            
            & 3 & {2} & \opti{-0.56} & \opti{0.} & \opti{0.17} & \opti{-0.56} & \opti{0.} & {500} & \opti{-0.56} & \opti{0.} & {6} & {-0.57} & {0.007} \\
            
            & 4 & \opti{146} & \opti{-0.76} & \opti{0.01} & {213} & {-0.78} & {0.04} & \textsc{oot} & \textsc{oot} & \textsc{oot} & {171} & {-0.73} & {0.03} \\
            
            \multirow{-5}{*}{\shortstack[C]{\textsc{Adversarial} \\ \textsc{Tiger}}} & 5 & \opti{2082} & \opti{-0.96} & \opti{0.02} & {3237} & \opti{-0.96} & {0.06} & \textsc{oot} & \textsc{oot} & \textsc{oot} & {2583} & \opti{-0.96} & \opti{0.02}\\

        \bottomrule
        \end{tabular}
        \vspace{-.5cm}
        }
\end{table}

\subsection{Implementation Details}

The results presented in this paper were run on CPU (AMD Ryzen 7 PRO).

\end{document}